\pdfoutput=1
\documentclass[10pt,a4paper]{article}

\usepackage[margin=1in]{geometry}
\usepackage{graphicx}
\usepackage{xcolor}
\usepackage[colorlinks=true,allcolors=blue]{hyperref}
\usepackage{amsmath}   % math environments
\usepackage{amssymb}   % \mathbb
\usepackage{natbib}    % \citep / \citet
\usepackage{booktabs}  % \toprule / \midrule / \bottomrule
\usepackage{multirow}  % \multirow in appendix tables
\usepackage{enumitem}  % [nosep] and [label=(\alph*)]
\usepackage{colortbl}  % restrained row emphasis in comparison tables
\usepackage{siunitx}   % decimal alignment in numeric tables
\usepackage{placeins}  % keep Results displays before Discussion
\definecolor{HapiRow}{HTML}{EAF3F8}
\newcommand{\hapirow}{\rowcolor{HapiRow}}
\usepackage{titlesec}
\titlespacing*{\paragraph}{0pt}{0.5em}{1em}
\newcommand{\orcid}[1]{\href{https://orcid.org/#1}{\includegraphics[width=8pt]{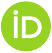}}}
\newcommand{\ack}[1]{\section*{Acknowledgments}#1}
\newcommand{\funding}[1]{\section*{Funding}#1}
\newcommand{\data}[1]{\section*{Data availability}#1}

\begin{document}

\title{Hapi: A Multivariable Land-Surface Transformer for Medium-Range Hydrological Forecasting at Continental Scale}

% ------------------------------------------------------------------
% AUTHOR BLOCK --- fill in per the ML:E author guidelines.
% Add verified \orcid{XXXX-XXXX-XXXX-XXXX} identifiers before submission;
% corresponding author marked $^{n,*}$.
% ------------------------------------------------------------------
\author{Hong Zhang$^{1,*}$\orcid{0000-0002-7720-0049}, John K. Hutchison$^{1}$\orcid{0000-0002-7419-6611}, Rao Kotamarthi$^{1}$\orcid{0000-0002-2612-7590}\\
Jeremy Feinstein$^{1}$\orcid{0009-0009-5362-6860}, Haiwen Guan$^{1}$\orcid{0009-0008-3005-1497}, Romit Maulik$^{1,2}$\orcid{0000-0002-4370-0416}\\
Ross M. Alexander$^{1}$\orcid{0000-0003-1106-1100}, Vijay P. Ramalingam$^{1}$\orcid{0009-0008-1599-5676}, Jason Stock$^{1}$\orcid{0009-0006-2354-2373}, Thomas Wall$^{1}$\orcid{0000-0002-4070-8623}\\[0.75em]
\small $^{1}$Argonne National Laboratory, Lemont, IL 60439, USA\\
\small $^{2}$Purdue University, West Lafayette, IN 47907, USA\\
\small $^{*}$Corresponding author: \href{mailto:hongzhang@anl.gov}{hongzhang@anl.gov}}
\date{}
\maketitle

\begin{abstract}
Accurate flood forecasts several days in advance are essential for flood control, water-resource management, and emergency response.
A central challenge is to produce high-resolution forecasts across continental domains where hydrological behavior varies widely from place to place.
We developed Hapi, a U-Net Swin Transformer that uses fine three-dimensional patches and hierarchical shifted-window attention to forecast river discharge, surface runoff, snow water equivalent, and soil wetness index across the contiguous United States.
The model produces medium-range forecasts (24--72~h) at $0.05^{\circ}$ resolution with adaptive task weighting and required only 0.11 seconds for a four-variable 72-h CONUS forecast on one A100 GPU.
In a held-out 2024 potential-skill evaluation with ERA5-Land inputs prescribed over the forecast horizon, Hapi achieved the highest F1-score for floods in 20 of 21 comparisons across seven GloFAS return periods and three forecast leads.
Independent validation against observed daily discharge at 3{,}881 U.S.\ Geological Survey gauges showed that Hapi achieved the highest median Nash--Sutcliffe efficiency at every lead, supported by regional-cluster bootstrap intervals.
In a matched 24-h comparison of loss formulations, adaptive task balancing produced the lowest discharge errors and the highest F1-score for floods.
\end{abstract}

\noindent\textbf{Keywords:} deep learning, hydrological forecasting, flood forecasting, land-surface modeling, Swin Transformer, continental-scale forecasting

\section{Introduction}
\label{sec:intro}

Accurate flood forecasts several days in advance are essential for flood control, water-resource management, and emergency response~\citep{tellman2021satellite,alfieri2013glofas}.
The central challenge is high resolution over continental domains, where hydrological behavior varies widely from place to place.
Operational systems address this challenge with process-based simulators.
The Global Flood Awareness System (GloFAS)~\citep{alfieri2013glofas}, and the U.S.\ National Water Model~\citep{cosgrove2024nwm}, for example, solve land-surface and river-routing equations at continental scale and produce a full land-surface state.
Their ensembles consume millions of CPU-hours per year, depend on hand-tuned regional parameters~\citep{mizukami2017seamless,beck2020global}, and take months to adapt to new variables or regions.

\begin{figure}[t]
    \centering
    \includegraphics[width=\linewidth]{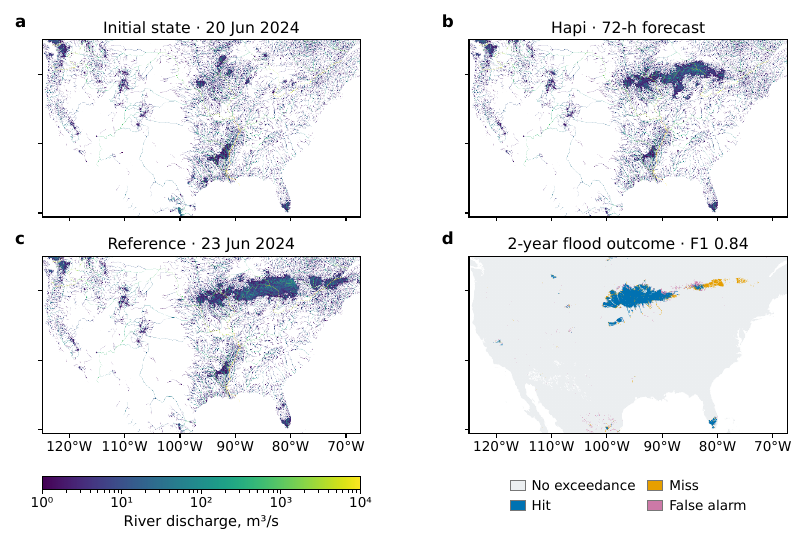}
    \caption{Continental illustration of a Hapi discharge forecast.
    \textbf{a}, Discharge at initialization on 20 June 2024, represented by
    the field carried forward in the persistence forecast. \textbf{b}, Hapi's 72-h discharge forecast valid on
    23 June. \textbf{c}, GloFAS reanalysis for the same valid date.
    \textbf{d}, Correctly predicted exceedances (hits), missed exceedances,
    and false alarms at each cell's local 2-year return level; cells with no
    forecast or reference exceedance are shown in grey.  The valid date was
    selected independently of Hapi as the 2024 test date with the largest
    number of reference 2-year-threshold exceedances over CONUS.  The
    discharge maps share a logarithmic colour scale.}
    \label{fig:conus_demo}
\end{figure}

Deep learning provides a complementary approach, but most hydrological applications have represented rivers as collections of individual basins or gauges.
For basin- and gauge-based prediction, long short-term memory networks (LSTMs) are widely used to predict streamflow~\citep{nearing2021role}.
In earlier work, regional LSTMs were trained across multiple catchments using static basin attributes~\citep{kratzert2019lstm,newman2015camels,addor2017camels}.
Regional LSTM applications now span multiple timescales, global regionalization, and operational flood warning~\citep{gauch2021multitimescale,nearing2024floods,nevo2022floodhub}.
Caravan~\citep{kratzert2023caravan} and NeuralHydrology~\citep{kratzert2022neuralhydrology} provide standardized datasets and workflows for building and evaluating these basin-based models.
Transformers and time-series foundation models have also been applied to 120-h streamflow prediction and zero-shot forecasting at unseen stations~\citep{demiray2026generalized,sun2026zeroshot}.
Despite their broader reach, these models produce basin or gauge sequences without explicit spatial interaction across a gridded domain.

Gridded models extend deep-learning hydrological prediction beyond this pointwise representation.
RiverMamba~\citep{shamseddin2025rivermamba} brings spatial context to global $0.05^{\circ}$ discharge forecasting and outperforms an operational LSTM baseline, particularly beyond 48~h.
It traverses the grid as a one-dimensional sequence with bidirectional Mamba blocks~\citep{gu2024mamba}, forming a serial scan order on a two-dimensional field.
SwinFlood~\citep{song2025swinflood} combines a convolutional neural network with Swin Transformer blocks within a single urban--rural catchment on the Shenzhen River.
These models move hydrological deep learning from independent basin sequences to spatially distributed prediction.

Related developments in weather and Earth-system forecasting offer strategies for handling the spatial dimension of gridded forecasting.
Hierarchical and graph-based models exchange information across large spatial domains without exhaustive pairwise interactions between all locations~\citep{bi2023pangu,lam2023graphcast,chen2023fuxi}.
Pangu-Weather~\citep{bi2023pangu} adds a 3D Earth-specific transformer with an explicit pressure-level axis, matched to atmospheric vertical structure but unnatural for surface-only variables.
GraphCast~\citep{lam2023graphcast} uses a multi-mesh icosahedral graph that trades resolution uniformity for global coverage.
FuXi~\citep{chen2023fuxi} and Aurora~\citep{bodnar2025aurora} use U-shaped vision transformers on a two-dimensional grid, but are designed for atmospheric targets.
The same architectural choices also affect the resolution a model can reach.
Since the cost of attention grows quadratically with the number of tokens, models using global attention are usually limited to coarse grids.
For example, Stormer~\citep{nguyen2024stormer}, a standard vision transformer, runs at $1.40625^{\circ}$; ClimaX~\citep{nguyen2023climax}, also global, runs at $1.40625^{\circ}$ and coarser.
All the window- and mesh-based models mentioned above run at $0.25^{\circ}$.
Aurora reaches $0.1^{\circ}$, but only by enlarging its patches (thus a lower token count).
Beyond spatial resolution, continental forecasting can involve predicting several hydrological variables together.
Generalist Earth-system models use shared architectures to forecast multiple fields~\citep{bodnar2025aurora,nguyen2023climax,schmude2024prithviwxc,vaughan2025aardvark}, and ecLand emulators advance several prognostic land-surface states together~\citep{wesselkamp2025ecland}.
Existing models address spatial or multivariable prediction, but efficient, high-resolution continental forecasting of multiple hydrological states remains underexplored.

To address this gap, we introduce \textbf{Hapi}, a hierarchical transformer that forecasts river discharge, surface runoff, snow water equivalent, and soil wetness over the contiguous United States (CONUS) out to 72~h.
Figure~\ref{fig:conus_demo} shows an example of Hapi's 72-h discharge forecast across CONUS.
Hapi combines fine three-dimensional patches with hierarchical shifted-window attention that expands spatial context without global attention over the full grid.
Learned task weights account for the different scales and distributions of the four targets.
During autoregressive rollout, the four predicted fields feed into the next step, while meteorological forcing is updated externally.
% The resulting fields share a grid and forecast schedule, which simplifies their integration with broader forecasting systems; this study does not evaluate such coupling.
On a 2024 CONUS holdout, we evaluate river discharge and flood forecasting against persistence, the operational GloFAS forecast, and RiverMamba.
ERA5-Land fields supply forcing and land-surface covariates over the forecast horizon.
The evaluation therefore measures Hapi's hydrological potential skill when these inputs are prescribed, while a precipitation-perturbation experiment probes sensitivity to forcing errors.
%We also measure Hapi's inference cost.
Our contributions are

\begin{itemize}[nosep]
    \item \textbf{An integrated system for high-resolution continental hydrology.}
    Hapi combines U-Net and Swin V2 components in an end-to-end model that preserves the two-dimensional grid and reconstructs full-resolution outputs, as described in Section~\ref{sec:method}.
    We demonstrate the system with a combination of $0.05^{\circ}$ continental coverage, four jointly predicted hydrological states, and 24--72-h rollout.

    \item \textbf{Automatic balancing of heterogeneous hydrological targets.}
    Hapi forecasts discharge, runoff, snow water equivalent, and soil wetness without manually tuning their relative loss weights, despite their different scales and error distributions, as described in Section~\ref{sec:loss}.
    In a matched 24-h comparison of three loss formulations, learned task weighting produced the lowest discharge errors and the highest F1-score for floods, with the largest F1 differences at longer return periods, as shown in Section~\ref{sec:ablation_loss}.

    \item \textbf{Leading flood forecasting with independent observational validation.}
    Hapi achieves the highest F1-score for floods in 20 of 21 lead--return-period comparisons in Section~\ref{sec:results_discharge}, including every return period from 1.5 to 100~yr at 48 and 72~h.
    All 21 F1 advantages over RiverMamba remain significant after multiplicity correction, as shown in Appendix~\ref{app:bootstrap_paired}.
    Hapi also has the highest base-rate-invariant symmetric extremal dependence index, SEDI, in 18 of 21 comparisons with GloFAS and persistence in Appendix~\ref{app:sedi}.
    Against observed daily discharge at 3{,}881 U.S.\ Geological Survey (USGS) gauges, Hapi also achieves higher median Nash--Sutcliffe efficiency (NSE) than GloFAS and persistence at every lead.
    Section~\ref{sec:gauge} presents this independent observational validation.

    % \item \textbf{Fast continental inference.}
    % Hapi produces a 72-h forecast of all four variables over CONUS in an average of 0.11\,s with 1.69\,GB of peak allocated memory on one A100 GPU, as reported in Section~\ref{sec:efficiency}.
\end{itemize}

\section{Data}
\label{sec:data}

We use a daily gridded dataset assembled from GloFAS reanalysis~\citep{harrigan2020glofas}, ERA5-Land~\citep{c3s2019era5land}, and static terrain and upstream drainage fields.
The dataset spans twenty years, 2005--2024, over CONUS on a $0.05^{\circ}$ latitude--longitude grid, approximately 5\,km, at a 24-h time step.
ERA5-Land data are regridded from their native $0.1^{\circ}$ resolution to the GloFAS grid.
The GloFAS grid consists of $512 \times 1152 = 589{,}824$ cells.
Ocean cells and cells outside CONUS are zero-padded to preserve the rectangular grid shape.
Because the source fields have different spatial coverage, each variable has its own valid-cell mask for loss calculation and autoregressive feedback, as described in Appendix~\ref{app:masking}.

\subsection{Variables}
\label{sec:variables}

Each sample from the dataset contains 12 input variables grouped by their role in the autoregressive rollout described in Section~\ref{sec:method}.
Table~\ref{tab:variables} lists their sources, units, and definitions.

\begin{table}[h]
\centering
\caption{The 12 model inputs, grouped by their role in the autoregressive rollout. ERA5-Land fields come from the Copernicus Climate Change Service Climate Data Store~\citep{c3s2019era5land}; upstream drainage area comes from the GloFAS v4.0 auxiliary data~\citep{ecmwf2023glofasaux}. DEM denotes digital elevation model.}
\label{tab:variables}
\small
\renewcommand{\arraystretch}{1.06}
\setlength{\tabcolsep}{5pt}
\begin{tabular}{@{}p{0.13\textwidth}p{0.18\textwidth}p{0.63\textwidth}@{}}
\toprule
\textbf{Role} & \textbf{Source} & \textbf{Variables and units} \\
\midrule
Prognostic & GloFAS reanalysis & River discharge, m$^3$\,s$^{-1}$; runoff water equivalent, mm; snow water equivalent, mm; soil wetness index, dimensionless \\
\midrule
Forcing & ERA5-Land & Total runoff, \texttt{ro}, m; precipitation, \texttt{tp}, m; evaporation, \texttt{e}, m; soil water in layer~1, \texttt{swvl1}, m$^3$\,m$^{-3}$; solar radiation, \texttt{ssrd}, J\,m$^{-2}$; 2\,m air temperature, \texttt{t2m}, K \\
\midrule
Static & DEM; GloFAS v4.0 & Surface elevation, m; upstream drainage area, m$^2$ \\
\bottomrule
\end{tabular}
\end{table}

ERA5-Land supplies the six forcing variables at each rollout step, making the prediction of hydrological states reanalysis-conditioned.
The two runoff variables differ because GloFAS runoff is routed and prognostic, whereas ERA5-Land \texttt{ro} is an unrouted forcing.
Similarly, \texttt{swvl1} describes the top 7\,cm, while the prognostic soil wetness index aggregates the top three layers with vegetation-dependent weights.
% ERA5-Land evaporation retains its native negative sign and can be replaced operationally by forecast evaporation or an atmospheric reference-evapotranspiration estimate.
Appendix~\ref{app:normalization} reports the variable-specific normalization choices and the distributions that motivate them.

\subsection{Data Splits}
\label{sec:splits}

We follow the WeatherBench convention and split the data by calendar year.
Models are trained on 2005--2022, comprising eighteen years and approximately 6{,}575 daily samples, while the 365 samples from 2023 are used only for validation.
The held-out test data window runs from 1 January through 1 November 2024.
After the $T{=}2$ history buffer and three-step rollout are accounted for, it provides 302 forecasts at each lead and is used only for the results in Section~\ref{sec:experiments}.
The three calendar ranges are disjoint and ordered in time.

\subsection{Flood Thresholds}
\label{sec:thresholds}

Following RiverMamba~\citep{shamseddin2025rivermamba}, we use the per-cell return levels distributed as GloFAS v4.0 auxiliary data to identify floods at different return periods~\citep{ecmwf2023glofasaux}.
GloFAS derives these fields by fitting a Gumbel distribution, the $\xi = 0$ member of the generalized extreme value family, to annual maxima using L-moments.
The annual maxima come from LISFLOOD~\citep{burek2013lisflood} physical simulations forced with ERA5 in the GloFAS reanalysis~\citep{harrigan2020glofas} for 1979--2022.
The fitted distributions provide return levels for 1.5, 2, 5, 10, 20, 50, 100, 200, and 500 years.
At each grid cell $\mathbf{x}$, the $r$-year return level $q_r(\mathbf{x})$ is the discharge threshold satisfying $\Pr[Q_{\max}(\mathbf{x})>q_r(\mathbf{x})]=1/r$, where $Q_{\max}(\mathbf{x})$ is the annual maximum discharge. Thus, a 100-year return level has a 1\% chance of being exceeded in any given year, and exceedances can occur in consecutive years.
Using these published fields keeps the event definition consistent with the one used by GloFAS for flood warnings.
The fitting window ends in 2022, before the 2024 test window, so the thresholds carry no information from the validation or test periods.

The return-level fields align exactly with our grid, allowing us to select a CONUS subset without interpolation, as described in Appendix~\ref{app:masking}.
The intersection of defined return levels and the discharge mask contains the $414{,}795$ cells used for discharge evaluation.

For each return period $r$, a forecast or reference value is classified as a flood when its discharge reaches or exceeds the corresponding $r$-year return level.
The same return level is applied to every model and the GloFAS reanalysis reference.
% The same thresholds are applied to all baselines, making the F1, precision, and recall values in Table~\ref{tab:flood_f1} directly comparable across models.

\section{Method}
\label{sec:method}

\begin{figure}[t]
    \centering
    \includegraphics[width=\linewidth]{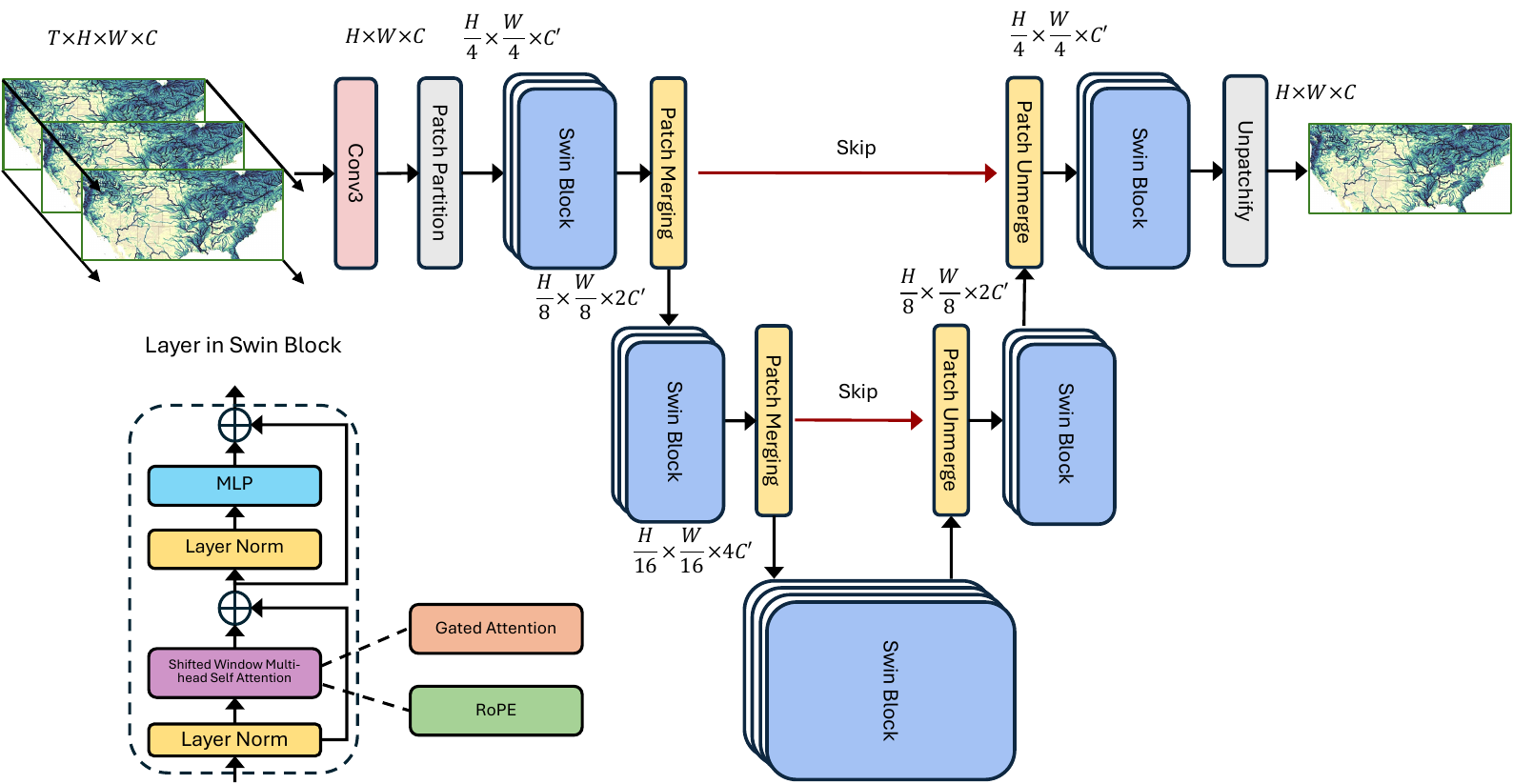}
    \caption{Hapi architecture.
    The encoder on the left downsamples through $N$ hierarchical Swin stages, and the symmetric decoder on the right mirrors the first $N-1$ stages after the shared bottleneck.
    Each Swin block uses 2D rotary position embeddings and gated attention, both described below, while skip connections transfer fine-grained features from matching encoder stages to the decoder.
    During autoregressive rollout, prognostic channels are predicted and fed back, forcing channels are supplied externally, and static channels remain fixed, as described in Section~\ref{sec:rollout}.}
    \label{fig:architecture}
\end{figure}

Hapi is a U-Net Swin Transformer V2 for gridded multivariable hydrology at $0.05^{\circ}$ resolution.
Its backbone integrates a U-Net~\citep{ronneberger2015unet} encoder--decoder, Swin Transformer V2 blocks~\citep{liu2022swinv2}, 2D rotary position embeddings~\citep{su2024roformer,fang2023eva02}, and gated attention~\citep{qiu2025gatedattention}.
Given $T$ historical frames of $C_{\mathrm{in}}$ input variables on an $H \times W$ grid, the model predicts $C_{\mathrm{out}}$ prognostic variables at the next time step, as shown in Figure~\ref{fig:architecture}.

Gridded forecasting models differ in how they convert input data into tokens, using per-variable tokenization with cross-attention~\citep{nguyen2023climax}, physically grouped branches~\citep{bi2023pangu}, or variable-agnostic latent encoders~\citep{jaegle2022perceiverio,bodnar2025aurora}.
Hapi keeps a monolithic patch embedding because every input shares one grid and one set of history frames and does not need cross-resolution or cross-modality alignment.
At the rollout level, Hapi separately treats prognostic, forcing, and static inputs according to their roles.

\subsection{Architecture}

To limit early spatial aggregation during tokenization, Hapi uses small $p\times p$ spatial patches with $p=2$.
A three-dimensional convolutional (Conv3D) layer, with kernel and stride $(T, p, p)$ maps the input tensor $\mathbf{X} \in \mathbb{R}^{B \times C_{\mathrm{in}} \times T \times H \times W}$ to $\frac{H}{p} \times \frac{W}{p}$ tokens of dimension $d$, where $B$ is the batch size.
This projection combines the variables and $T$ historical frames within each patch, leaving a two-dimensional token grid for subsequent attention.

Hapi uses $N$ stages of Swin Transformer blocks to expand its receptive field without paying the cost of global attention that a conventional vision transformer requires.
Attention operates within local windows, and shifted windows allow information to pass across window boundaries.
Between stages, patch merging halves each spatial dimension and doubles the channel width.
This encoder expands the area covered by each window while reducing the number of tokens at the next stage.

The decoder mirrors the encoder's structure by upsampling through patch unmerging, which uses a learned linear expansion and pixel-shuffle rearrangement to double each spatial dimension.
Skip connections concatenate encoder and decoder features at corresponding resolutions, giving the decoder direct access to encoder representations that bypass the bottleneck.
A final linear projection produces $C_{\mathrm{out}}$ channels at the original $H \times W$ resolution, one predicted field per prognostic variable.

\paragraph{Position encoding.}
\label{sec:rope}

Within an attention window, token interactions depend on relative spatial position as well as the local features.
Swin V2 represents position through a learned multilayer perceptron (MLP), that maps coordinate offsets $(\Delta h, \Delta w)$ to an additive attention bias.
We use axial 2D rotary position embeddings (RoPE)~\citep{su2024roformer,heo2024ropevit}, to encode these offsets directly in the query--key interaction without learning a position-bias network.
For each attention head of dimension $d_h$, we group vector components into even--odd pairs and alternate row and column rotations across successive pairs.
For a query vector $\mathbf{q}$ at window position $(h, w)$, the rotation is
\begin{equation}
    \mathrm{RoPE}(\mathbf{q}_{2k}, \mathbf{q}_{2k+1}; h, w) =
    \begin{pmatrix} \cos\phi_k & -\sin\phi_k \\ \sin\phi_k & \cos\phi_k \end{pmatrix}
    \begin{pmatrix} \mathbf{q}_{2k} \\ \mathbf{q}_{2k+1} \end{pmatrix},
\end{equation}
where the angles are defined by
\begin{equation}
    \phi_{2j}(h,w) = h\,\omega_j, \qquad
    \phi_{2j+1}(h,w) = w\,\omega_j, \qquad
    \omega_j = \theta^{-4j/d_h},
\end{equation}
with $j = 0,\ldots,d_h/4-1$ indexing the frequency bands.
The coordinates $h$ and $w$ index the row and column within each attention window.
Applying the same rotations to key vectors makes the query--key dot product depend on position through relative row and column offsets.
We set $\theta = 10$ for the $8\times 8$ windows.

\paragraph{Gated attention.}
\label{sec:gated}

A Swin window can contain river channels, dry land, and ocean cells, so the relevance of neighboring information can vary across locations and features.
Within each head, standard softmax attention uses one normalized set of weights to mix all features of the value vectors.
Consequently, it passes a mixture of values forward, even when none provides useful information for a particular feature of the receiving token.
We add separate, query-dependent control using the element-wise sigmoid gate of \citet{qiu2025gatedattention}, computed from the receiving token and applied after attention.
The gate allows each token to attenuate individual attended features before the output projection.
For input token features $\mathbf{X}$ within a window, we compute the gate logits $\mathbf{G}$ together with the query, key, and value projections:
\begin{equation}
    [\mathbf{Q},\mathbf{K},\mathbf{V},\mathbf{G}]
    = \mathbf{X}\mathbf{W}_{qkvg}+\mathbf{b}_{qkvg}.
\end{equation}
The gated attention output is
\begin{equation}
    \mathbf{O} = \sigma(\mathbf{G}) \odot
    \mathrm{Attn}(\mathbf{Q},\mathbf{K},\mathbf{V}),
\end{equation}
where $\sigma$ is the sigmoid function and $\odot$ denotes element-wise multiplication.
The gate $\sigma(\mathbf{G})$ assigns a scale between zero and one to each feature in every token and attention head.

\subsection{Autoregressive Rollout with Prognostic--Forcing Separation}
\label{sec:rollout}

The model rolls out by feeding its output back as input.
Variables update by different rules at each step: \textbf{prognostic} variables, comprising river discharge, runoff, snow water equivalent, and soil wetness index, are replaced by their predicted fields; \textbf{forcing} variables, including precipitation, temperature, and radiation, are supplied from ERA5-Land; \textbf{static} descriptors, comprising elevation and upstream drainage area, stay fixed.
This separates the prognostic hydrological state from meteorological forcing, so forecast products can supply the forcing fields while terrain remains fixed at every step.
This prognostic--forcing separation, with prescribed drivers supplied throughout the rollout, is also used in land-surface emulation~\citep{wesselkamp2025ecland}.
Prognostic predictions are held within declared physical ranges before they are fed back.
The nonnegative variables are snapped to the point mass at zero in their target distributions, and the soil wetness index is clipped to $[0,1]$, as described in Appendix~\ref{app:denoising}.
The rule is applied only at intermediate steps and is identical during training and inference; the loss at the final step is computed from the raw output.
% AI weather models stabilize long horizons with curriculum fine-tuning~\citep{lam2023graphcast}, randomized forecast intervals~\citep{nguyen2024stormer}, or diffusion-based forecasting~\citep{price2025gencast,hatanpaa2025aeris}.
To stabilize long horizons, we fine-tune on rollouts of randomized depth, in the manner of \citet{brandstetter2022message}, under the training schedule described in Section~\ref{sec:loss}.

\subsection{Training Objective}
\label{sec:loss}

Foundation models such as Pangu-Weather~\citep{bi2023pangu}, GraphCast~\citep{lam2023graphcast}, FuXi~\citep{chen2023fuxi}, and GenCast~\citep{price2025gencast} balance per-variable losses with hand-tuned scalar weights.
RiverMamba instead uses fixed severity and lead-time weights~\citep{shamseddin2025rivermamba}, but these do not balance variables because RiverMamba predicts only discharge.
For Hapi's four targets, we adapt the likelihood-based Gaussian task weighting of \citet{kendall2018multi}, which uses mean squared error, to a Laplacian form with mean absolute error (MAE) for hydrology's heavy-tailed targets with frequent zeros.
Per-variable log-scales are learned end to end.
Loss is computed only on the $C_p$ prognostic outputs, so $C_{\mathrm{out}} = C_p$.

The prognostic targets span six orders of magnitude, and up to 78\% of their values are exactly zero, as shown in Table~\ref{tab:normalization}, making uniform standardization unsuitable.
Discharge, snow water equivalent, and runoff are reparameterized with $\log(1+x)$ so that rare large values do not dominate the loss, as described in Appendix~\ref{app:normalization}.
For a prediction $\hat{\mathbf{Y}} \in \mathbb{R}^{B \times C_p \times H \times W}$ and target $\mathbf{Y}$, define the residual $\delta_{b,i,h,w} = \hat{Y}_{b,i,h,w} - Y_{b,i,h,w}$ in normalized space.
The masked mean absolute error for variable $i$ is
\begin{equation}
    \mathrm{MAE}_i = \frac{\sum_{b,h,w} m_{i,h,w} \, \lvert \delta_{b,i,h,w} \rvert}
               {B \sum_{h,w} m_{i,h,w}},
    \qquad i = 1, \dots, C_p,
\end{equation}
where $m_{i} \in \{0,1\}^{H \times W}$ is the per-variable validity mask of Appendix~\ref{app:masking}, so zero-padded ocean and out-of-domain cells neither contribute to the numerator nor inflate the denominator.

Each variable has a learnable log-scale $s_i$.
Writing $\delta_i$ for a scalar residual of variable $i$, the Gaussian formulation of \citet{kendall2018multi} gives the negative log-likelihood $\tfrac{1}{2}e^{-2s_i}\delta_i^2 + s_i$ up to an additive constant, with $s_i$ denoting the log-standard deviation.
For our MAE objective, we use a zero-centered Laplacian residual model with scale $e^{s_i}$:
\begin{equation}
    p(\delta_i \mid s_i)
    = \frac{1}{2} e^{-s_i}\,
      \exp\!\left[- e^{-s_i}\,\lvert\delta_i\rvert\right].
\end{equation}
Averaging its negative log-likelihood over valid cells and the batch, and dropping the constant $\log 2$, gives $e ^{-s_i}\,\mathrm{MAE}_i + s_i$.
Averaging over variables yields
\begin{equation}
    \mathcal{L} = \frac{1}{C_p} \sum_{i=1}^{C_p} \bigl( e^{-s_i}\,\mathrm{MAE}_i + s_i \bigr),
    \label{eq:loss_mae}
\end{equation}
where $e^{-s_i}$, the inverse Laplace scale, is the learned task weight.
Increasing $s_i$ reduces this weight, and the additive $s_i$ term prevents all weights from collapsing to zero.
For fixed $\mathrm{MAE}_i>0$, stationarity with respect to $s_i$ gives $e^{-s_i} = 1/\mathrm{MAE}_i$.
The learned weights therefore balance the variables according to their error scales, removing the need to tune their relative loss weights by hand.
The $s_i$ are global task-level scales shared across samples, locations, and lead times.
They adapt the relative contributions of the four prognostic variables during optimization; they are not input-dependent predictive uncertainties and do not provide calibrated forecast intervals.
Appendix~\ref{app:implementation} reports the converged task log-scales $\{s_i\}$.

Multi-step fine-tuning follows \citet{brandstetter2022message} by randomizing the rollout horizon.
At each iteration, we sample $J \sim \mathrm{Uniform}\{1,\dots,J_{\max}\}$, autoregressively roll out the model for $J$ steps, and evaluate the loss on the prediction at step $J$:
\begin{equation}
    \mathcal{L}_{\mathrm{total}}
    = \mathbb{E}_{J \sim \mathrm{Uniform}\{1,\dots,J_{\max}\}}
      \left[\mathcal{L}^{(J)}\right].
\end{equation}
The intermediate predictions are fed back into the model without detaching the computational graph, so gradients from $\mathcal{L}^{(J)}$ propagate through the complete $J$-step rollout.
We use $J_{\max}{=}3$.

\subsection{Final Model Specification}

The final model uses two consecutive daily frames, $T{=}2$, of 12 input variables to predict the next-day fields of four prognostic variables.
Longer histories produced only marginal accuracy gains at greater computational cost.
Although the temporal projection can be reconfigured for any history length, $T{=}2$ provided the best accuracy--efficiency trade-off among those tested.

The three encoder stages contain $(1,2,4)$ Swin blocks, and the decoder mirrors the first two stages with $(2,1)$ blocks after the shared bottleneck.
Thus, the full symmetric encoder--decoder path has block depths $(1,2,4,2,1)$.
The model uses embedding dimension $d{=}192$, patch size $p{=}2$, and window size $w{=}8$.
The complete model has 43.96 million parameters.
These four hyperparameters were selected through Bayesian optimization of validation loss.
Appendix~\ref{app:sweep} reports the search space and results.

Training begins with single-step pretraining and continues with randomized multi-step fine-tuning using $J_{\max}{=}3$.
We optimize the task log-scales $\{s_i\}$ during pretraining and freeze them during fine-tuning.
Freezing the scales prevents them from compensating for accumulated rollout error, leaving the model parameters to reduce that error, as described in Appendix~\ref{app:twophase}.
Appendix~\ref{app:implementation} gives the full training details and learned log-scales.

\section{Results}
\label{sec:experiments}

We evaluated Hapi from overall forecast skill to specific capabilities and limitations.
We first compared discharge forecasts with persistence, GloFAS, and RiverMamba at continental and gauge scales.
We then compared four-variable and discharge-only prediction, tested three loss-weighting formulations, evaluated the other prognostic variables, and examined variation across river subsets and perturbed meteorological forcing.
Finally, we used Hurricane Helene to illustrate event-scale behavior and assessed computational performance.

\subsection{Baselines, Metrics, and Statistical Analysis}
\label{sec:baselines}

We evaluated gridded discharge against GloFAS reanalysis using a common verification target and compared Hapi with persistence, the operational GloFAS forecast, and RiverMamba.
The comparisons cover 24-, 48-, and 72-h leads during the 2024 test period.
Flood forecasting is the primary outcome, complemented by continuous discharge metrics, per-cell analyses, and paired moving-block bootstrap inference.
The verification target is the GloFAS reanalysis 24-h mean river discharge, the field Hapi is trained to predict in Section~\ref{sec:variables} and from which the per-cell return-level thresholds are derived in Section~\ref{sec:thresholds}.
Scores therefore measure agreement with that reanalysis, using an identical target for every model.

We compare Hapi against three baselines.
\textbf{Persistence} carries forward the last discharge field observed at forecast issue time and holds it constant across all leads; the 72-h forecast therefore uses the field from three days before the target.
The lag-1 autocorrelation of daily streamflow exceeds 0.95 on large catchments, making persistence a strong short-lead baseline.
\textbf{GloFAS forecast} is the operational, physics-based global flood forecasting system of the Copernicus Emergency Management Service~\citep{alfieri2013glofas}.
The GloFAS v4 forecasts evaluated here are generated with the distributed hydrological model LISFLOOD OS~\citep{burek2013lisflood}.
We use the Copernicus Emergency Management Service product \texttt{cems-glofas-forecast}, with system version \texttt{operational}, hydrological model LISFLOOD, the \texttt{control\_forecast} member, and variable \texttt{river\_discharge\_in\_the\_last\_24\_hours}, at lead times of 24, 48, and 72~h, retrieved over the same CONUS box as the training grid.
This operational forecast is evaluated against the same verification target as the other models.

\textbf{RiverMamba}~\citep{shamseddin2025rivermamba} is a recent gridded deep-learning discharge forecaster at the same $0.05^{\circ}$ resolution.
RiverMamba uses seven historical days through $t-1$, issues at $t$, and predicts $t+1$ through $t+7$, without using a day-$t$ discharge nowcast.
We follow this convention.
Thus the tabulated 24-, 48-, and 72-h leads follow each system's stated forecast issue time, although RiverMamba's latest discharge input is one day older than Hapi's at the same nominal lead.
Only Hapi and GloFAS carry the four prognostic land-surface variables; RiverMamba predicts discharge only.
RiverMamba was run on its native global domain for forecast evaluation, and its CONUS predictions were extracted for the comparisons below.

Each system is evaluated with the inputs it requires.
Hapi uses ERA5-Land forcing over the forecast horizon, whereas the operational GloFAS product uses forecast forcing.
RiverMamba uses an earlier discharge initialization, as described above.
These systems require different types of forecast input, so a single forcing dataset cannot be applied across them.
Nevertheless, the comparison evaluates their complete forecast performance against a common reference.

We exclude the U.S.\ National Water Model because its streamflow predictions are defined on NHDPlus river reaches rather than the regular grid used here~\citep{cosgrove2024nwm}.
A direct comparison would require a separately validated reach-to-grid mapping or an independent gauge evaluation that includes the National Water Model, both outside the scope of this study.

Continuous performance is evaluated using MAE and root mean squared error (RMSE), in m$^3$\,s$^{-1}$, together with the dimensionless NSE and Kling--Gupta efficiency, KGE.
The primary flood forecasting analysis uses per-cell GloFAS v4.0 return levels at seven return periods from 1.5 to 100~yr and reports F1-score, precision, recall, and the critical success index (CSI) for floods.
Exploratory results for the more uncertain 200- and 500-yr levels are reported in Appendix~\ref{app:extreme_tail}.
Definitions and aggregation conventions for all evaluation metrics are given in Appendix~\ref{app:metric_definitions}.
The primary discharge metrics use the 2024 test window at 24-, 48-, and 72-h leads on the 414{,}795-cell discharge mask.

The F1-score for floods is the primary outcome, and continuous discharge skill is secondary.
The decision a flood forecast supports is binary: at a given return period, warn or do not warn.
Threshold-based metrics directly quantify this decision, whereas continuous error statistics do not.
A forecast can have low RMSE while misclassifying threshold exceedances during flood events.
Unadjusted accuracy is uninformative at these base rates because a forecast with no warnings would score close to one.
F1 balances missed events against false alarms without requiring a cost ratio, which depends on the application.

F1 depends on the event base rate, precluding direct comparison between values at different return periods, such as 1.5 and 100~yr.
We use F1 to compare models at a fixed return period and the base-rate-invariant SEDI to compare skill across return periods, as described in Appendix~\ref{app:sedi}.
Pooled continuous metrics characterize domain-wide performance and emphasize high-discharge rivers.
We therefore also compute the metrics independently at each grid cell and summarize their spatial distributions in Appendix~\ref{app:percell}.

Uncertainty in F1 and SEDI differences is estimated with a paired moving-block bootstrap using 30-day blocks and $R=10{,}000$ replicates, as described in Appendix~\ref{app:bootstrap_paired}.
Within each replicate, the same days are resampled for both models.
Holm correction is applied across the 21 lead--return-period comparisons within each model pair and metric.
Hapi--persistence and Hapi--GloFAS comparisons use 302 common dates per lead, whereas Hapi--RiverMamba comparisons use the 294 dates available for both models.
We report paired differences with percentile intervals and multiplicity-adjusted decisions.

\subsection{Discharge Forecast Skill}
\label{sec:results_discharge}

\begin{figure}[t]
    \centering
    \includegraphics[width=\linewidth]{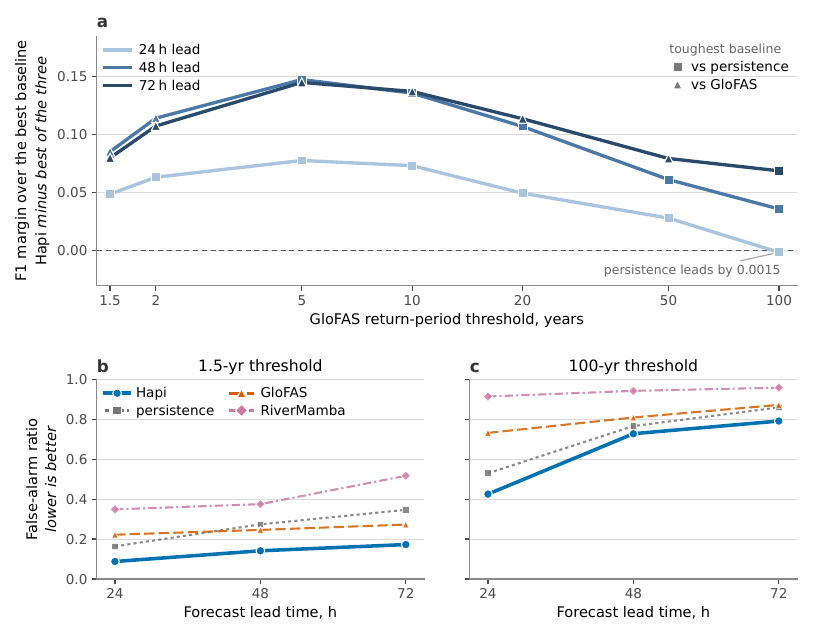}
    \caption{F1-score difference and false-alarm ratio for floods on the CONUS land mask in the 2024 test window.
    Panel a shows Hapi's F1 minus the highest F1 among the three baselines at each lead and return-period threshold.
    Line color denotes forecast lead; marker shape identifies the strongest baseline at each point.
    Panels b and c show the false-alarm ratio, $\frac{\mathrm{FP}}{\mathrm{TP}+\mathrm{FP}}$, at the 1.5-yr and 100-yr thresholds, respectively.
    Color, line style, and marker encode the model, with Hapi shown by thicker lines.}
    \label{fig:f1_curves}
\end{figure}

Hapi achieves the highest F1-score for floods in 20 of the 21 lead--return-period comparisons in Table~\ref{tab:flood_f1}, including all seven return periods from 1.5 to 100~yr at both 48 and 72~h.
Hapi exceeds RiverMamba throughout and falls below persistence only at the 100-yr threshold at 24~h, by 0.0015.
Figure~\ref{fig:f1_curves}a shows that Hapi's margin over the strongest baseline is larger at 48 and 72~h than at 24~h across all seven thresholds.
At the 1.5-yr and 100-yr thresholds shown in Figure~\ref{fig:f1_curves}b,c, Hapi has the lowest false-alarm ratio among the four models at every lead.

After Holm--Bonferroni correction, Hapi's F1 advantage over RiverMamba was significant in all 21 lead--threshold comparisons, and its advantage over GloFAS was significant at every threshold at 24~h.
For example, at the 5-yr threshold, the Hapi--GloFAS difference was $+0.203$ at 24~h (95\% interval, $0.165$--$0.227$) and $+0.145$ at 72~h (95\% interval, $0.087$--$0.182$).
Both differences remained significant after correction.

Among Hapi, GloFAS, and persistence, Hapi has the highest SEDI in 18 of 21 lead--threshold comparisons, including all seven at both 48 and 72~h.
The SEDI analysis in Appendix~\ref{app:sedi} complements the fixed-threshold F1 comparisons by allowing skill to be compared across event rarity.
At 72~h, Hapi's SEDI decreases from 0.931 at the 1.5-yr return period to 0.724 at the 100-yr return period; because SEDI is base-rate invariant, this decrease indicates lower skill for rarer floods.

In Table~\ref{tab:main_discharge}, Hapi has the lowest pooled MAE and RMSE at 24~h, but GloFAS leads on RMSE, NSE, and KGE at 48 and 72~h.
At 48~h, Hapi and GloFAS have nearly identical MAE, both rounding to 2.94\,m$^3$\,s$^{-1}$.
GloFAS has the lowest MAE at 72~h, while persistence has the highest KGE at 24~h.
Pooling emphasizes high-discharge rivers in a domain spanning five orders of magnitude of discharge, motivating a separate evaluation at individual grid cells.

Although GloFAS has higher pooled NSE at 48 and 72~h, Hapi has higher median per-cell NSE at both leads in Table~\ref{tab:percell}.
Among Hapi, GloFAS, and persistence, Hapi has the highest median per-cell NSE at every lead: 0.902, 0.823, and 0.776, compared with 0.698, 0.637, and 0.559 for GloFAS.
At 72~h, Hapi exceeds GloFAS at 79.2\% of scored cells and persistence at 92.3\%, with median NSE differences of $+0.142$ and $+0.363$, respectively.

\begin{table}[t]
\centering
\caption{F1-score for floods on the CONUS land mask in the 2024 test window at 24-, 48-, and 72-h leads and GloFAS return periods from 1.5 to 100~yr.
Higher is better; bold denotes the best score for each lead and threshold.
Results at the more uncertain 200- and 500-yr thresholds appear in Appendix~\ref{app:extreme_tail}.}
\label{tab:flood_f1}
\small
\setlength{\tabcolsep}{4pt}
\begin{tabular}{@{}llccccccc@{}}
\toprule
& & \multicolumn{7}{c}{\textbf{Return period, years}} \\
\cmidrule(lr){3-9}
\textbf{Model} & \textbf{Lead} & 1.5 & 2 & 5 & 10 & 20 & 50 & 100 \\
\midrule
persistence  & 24\,h & 0.836 & 0.799 & 0.709 & 0.627 & 0.562 & 0.496 & \textbf{0.470} \\
GloFAS       & 24\,h & 0.773 & 0.718 & 0.584 & 0.473 & 0.396 & 0.306 & 0.264 \\
RiverMamba   & 24\,h & 0.728 & 0.660 & 0.479 & 0.347 & 0.259 & 0.187 & 0.145 \\
\hapirow Hapi & 24\,h & \textbf{0.885} & \textbf{0.863} & \textbf{0.786} & \textbf{0.700} & \textbf{0.612} & \textbf{0.524} & 0.468 \\
\midrule
persistence  & 48\,h & 0.726 & 0.671 & 0.544 & 0.434 & 0.353 & 0.275 & 0.233 \\
GloFAS       & 48\,h & 0.747 & 0.685 & 0.535 & 0.413 & 0.333 & 0.239 & 0.191 \\
RiverMamba   & 48\,h & 0.697 & 0.622 & 0.433 & 0.301 & 0.212 & 0.138 & 0.098 \\
\hapirow Hapi & 48\,h & \textbf{0.833} & \textbf{0.799} & \textbf{0.691} & \textbf{0.569} & \textbf{0.460} & \textbf{0.337} & \textbf{0.269} \\
\midrule
persistence  & 72\,h & 0.653 & 0.590 & 0.447 & 0.334 & 0.256 & 0.177 & 0.140 \\
GloFAS       & 72\,h & 0.720 & 0.653 & 0.493 & 0.365 & 0.277 & 0.184 & 0.134 \\
RiverMamba   & 72\,h & 0.592 & 0.566 & 0.381 & 0.252 & 0.171 & 0.103 & 0.071 \\
\hapirow Hapi & 72\,h & \textbf{0.800} & \textbf{0.760} & \textbf{0.638} & \textbf{0.502} & \textbf{0.391} & \textbf{0.263} & \textbf{0.208} \\
\bottomrule
\end{tabular}
\end{table}

\begin{table}[t]
\centering
\caption{Discharge forecast skill on the CONUS land mask in the 2024 test window, pooled over space and time.
MAE and RMSE are in m$^3$\,s$^{-1}$; NSE and KGE are dimensionless.
Bold denotes the best score per column.
Per-cell skill is reported in Table~\ref{tab:percell}.}
\label{tab:main_discharge}
\small
\setlength{\tabcolsep}{3.7pt}
\begin{tabular}{@{}lcccccccccccc@{}}
\toprule
& \multicolumn{4}{c}{\textbf{24\,h lead}} & \multicolumn{4}{c}{\textbf{48\,h lead}} & \multicolumn{4}{c}{\textbf{72\,h lead}} \\
\cmidrule(lr){2-5} \cmidrule(lr){6-9} \cmidrule(lr){10-13}
\textbf{Model} & MAE & RMSE & NSE & KGE & MAE & RMSE & NSE & KGE & MAE & RMSE & NSE & KGE \\
\midrule
persistence            & 2.43 & 41.64 & 0.996 & \textbf{0.998} & 4.55 & 78.38 & 0.985 & 0.992 & 6.30 & 109.68 & 0.970 & 0.985 \\
GloFAS                 & 2.79 & 43.29 & 0.995 & 0.997 & \textbf{2.94} & \textbf{44.59} & \textbf{0.995} & \textbf{0.997} & \textbf{3.15} & \textbf{46.90} & \textbf{0.995} & \textbf{0.997} \\
RiverMamba             & 2.11 & 35.67 & 0.997 & 0.997 & 3.05 & 54.53 & 0.993 & 0.993 & 3.80 & 70.46 & 0.988 & 0.986 \\
\midrule
\hapirow Hapi          & \textbf{1.57} & \textbf{32.30} & \textbf{0.997} & 0.990 & \textbf{2.94} & 63.19 & 0.990 & 0.980 & 4.08 & 89.72 & 0.980 & 0.976 \\
\bottomrule
\end{tabular}
\end{table}

% Generated by paper/derive_percell_table.py -- do not edit by hand.
\begin{table}[t]
\centering
\caption{Per-cell continuous skill on the CONUS land mask in the 2024 test window. NSE, KGE, and the KGE bias ratio $\beta$ are computed independently at every grid cell, and their \emph{medians over cells} are reported, complementing the values pooled over space and time in Table~\ref{tab:main_discharge}. The last two columns give the percentage of cells at which Hapi's NSE exceeds that of the comparison model and the median per-cell difference. Scores use 374{,}297 cells, representing 90.2\% of the mask, where all three sources are defined at the relevant lead; the remainder have near-zero target variance, which makes NSE undefined. Bold marks the best NSE and KGE at each lead.}
\label{tab:percell}
\small
\setlength{\tabcolsep}{5pt}
\begin{tabular}{@{}llcccrc@{}}
\toprule
\textbf{Lead} & \textbf{Model} & Median NSE & Median KGE & Median $\beta$ & \multicolumn{2}{c}{Hapi wins vs.} \\
\cmidrule(lr){6-7}
 & & & & & \% cells & median $\Delta$NSE \\
\midrule
\hapirow 24\,h & Hapi & \textbf{0.902} & 0.891 & 0.973 & -- & -- \\
 & persistence & 0.823 & \textbf{0.911} & 1.000 & 89.4\% & +0.054 \\
 & GloFAS & 0.698 & 0.804 & 0.993 & 93.0\% & +0.142 \\
\addlinespace[3pt]
\hapirow 48\,h & Hapi & \textbf{0.823} & \textbf{0.832} & 0.967 & -- & -- \\
 & persistence & 0.560 & 0.779 & 1.000 & 91.5\% & +0.217 \\
 & GloFAS & 0.637 & 0.758 & 0.990 & 82.9\% & +0.124 \\
\addlinespace[3pt]
\hapirow 72\,h & Hapi & \textbf{0.776} & \textbf{0.795} & 0.962 & -- & -- \\
 & persistence & 0.339 & 0.669 & 1.000 & 92.3\% & +0.363 \\
 & GloFAS & 0.559 & 0.709 & 0.989 & 79.2\% & +0.142 \\
\bottomrule
\end{tabular}
\end{table}

\subsection{Independent Validation Against USGS Gauges}
\label{sec:gauge}

The preceding grid-based comparisons use GloFAS reanalysis as the verification target.
As an independent evaluation, we compare Hapi, GloFAS, and persistence against USGS daily discharge observations.
This comparison covers 3{,}881 gauges matched by drainage area to unique grid cells and retained with paired days during the 2024 test window.
The matching and filtering protocol is given in Appendix~\ref{app:gauge_protocol}.

Hapi has the highest median NSE at every lead, exceeding GloFAS at 67.1\%, 65.6\%, and 66.6\% of the 3{,}869 gauges with finite paired NSE at 24, 48, and 72~h, respectively (Table~\ref{tab:gauge_validation}).
Across the 3{,}863 paired gauges assigned to 18 two-digit hydrologic regions, Hapi's median NSE exceeded GloFAS's by 0.085--0.120 across the three leads.
The 95\% HUC2-cluster bootstrap intervals remained above zero at every lead, as detailed in Appendix~\ref{app:gauge_protocol}.

\begin{figure}[!htb]
\centering
\includegraphics[width=\textwidth]{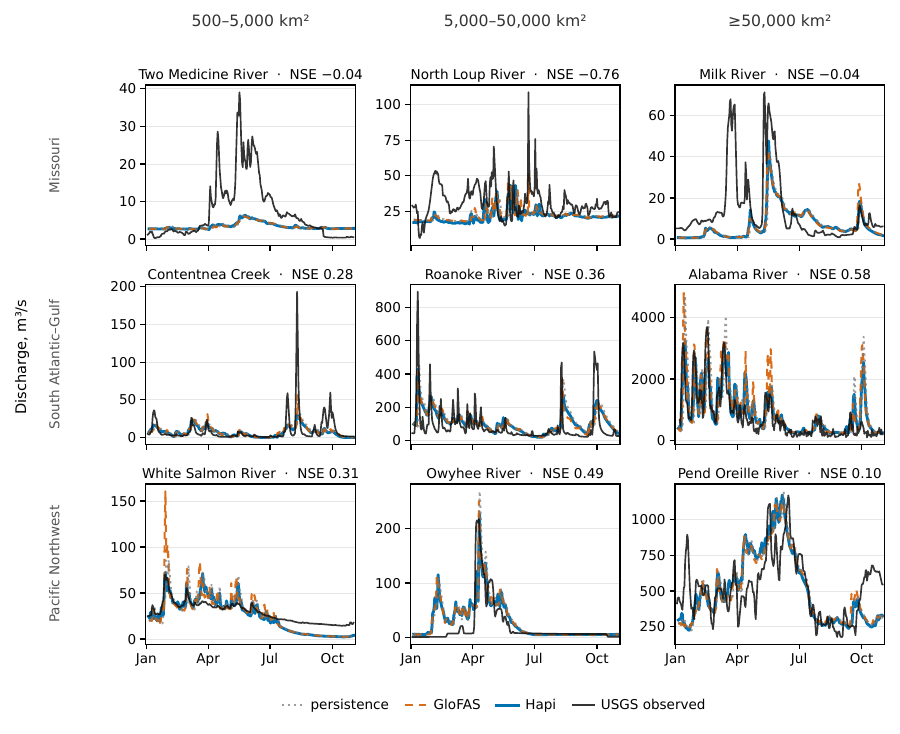}
\caption{Discharge at 72-h lead against USGS observations during the 2024 test window.
Rows show Missouri, South Atlantic--Gulf, and Pacific Northwest water-resource regions; columns show drainage areas of 500--5{,}000, 5{,}000--50{,}000, and at least 50{,}000~km$^2$.
Within each bin, the selected gauge has Hapi NSE nearest the bin median among gauges with at least 300 paired days and finite NSE.
The Alabama River bin contains only one eligible gauge.
Panel titles give Hapi NSE against observations; selection details are in Appendix~\ref{app:gauge_protocol}.}
\label{fig:hydrographs}
\end{figure}

Hapi's median NSE remains near 0.16 through 72~h, while that of persistence declines from 0.070 to $-0.108$.
Its median percent bias is also smaller, ranging from $+0.5$ to $+1.2$\%, compared with $+4.3$ to $+4.8$\% for GloFAS and $+5.3$ to $+5.5$\% for persistence.
Hapi therefore retains its advantage when evaluated against observed discharge, although performance varies substantially among gauges.
At 24~h, Hapi's median NSE is 0.160, with 10th and 90th percentiles of $-4.775$ and 0.661.
Its lowest median NSE occurs in basins larger than 50{,}000~km$^2$.
Figure~\ref{fig:hydrographs} illustrates this variation using gauges nearest the median Hapi 72-h NSE within each region and drainage-area band.
The Missouri panels show missed seasonal peaks on the Two Medicine and Milk rivers and muted variability on the North Loup River.
The South Atlantic--Gulf and Pacific Northwest panels generally track the observed variation more closely, though the Owyhee forecasts contain pulses with weak observational counterparts.
The hydrographs also demonstrate that a small median bias does not imply accurate discharge timing or magnitude at every gauge.
We next compared gauges whose station names contained terms associated with dams or reservoirs with the remaining gauges.
The 220 name-flagged gauges had a lower median 24-h NSE than the remaining gauges, 0.030 versus 0.168, with Mann--Whitney $p=0.004$.
Among basins larger than 50{,}000\,km$^2$, however, the difference is not statistically significant ($p=0.36$).

\begin{table}[!htb]
\centering
\caption{Independent validation against USGS daily discharge in the 2024 test window. Of 7{,}881 candidate gauges, 3{,}881 meet the matching and coverage criteria in Appendix~\ref{app:gauge_protocol}. NSE, KGE, RMSE, and percent bias, PBIAS, are medians across gauges. The final columns give the percentages with NSE above 0 and 0.5. Bold marks the best summary in each column.}
\label{tab:gauge_validation}
\small
\setlength{\tabcolsep}{3.5pt}
\begin{tabular}{@{}llcccccc@{}}
\toprule
\textbf{Lead} & \textbf{Model} & NSE & KGE & RMSE & PBIAS & $f_{\mathrm{NSE}>0}$ & $f_{\mathrm{NSE}>0.5}$ \\
 & & median & median & m$^3$/s & \% & \% & \% \\
\midrule
\hapirow 24\,h & Hapi & \textbf{0.160} & \textbf{0.310} & \textbf{19.4} & \textbf{+1.2} & \textbf{59.7} & \textbf{23.2} \\
 & GloFAS & 0.076 & 0.273 & 20.6 & +4.8 & 54.7 & 20.9 \\
 & persistence & 0.070 & 0.271 & 20.9 & +5.3 & 54.8 & 18.6 \\
\midrule
\hapirow 48\,h & Hapi & \textbf{0.165} & \textbf{0.312} & \textbf{19.3} & \textbf{+0.5} & \textbf{59.7} & \textbf{23.4} \\
 & GloFAS & 0.064 & 0.271 & 20.8 & +4.7 & 54.4 & 20.0 \\
 & persistence & -0.032 & 0.203 & 22.3 & +5.4 & 46.0 & 12.9 \\
\midrule
\hapirow 72\,h & Hapi & \textbf{0.159} & \textbf{0.306} & \textbf{19.4} & \textbf{+0.7} & \textbf{59.6} & \textbf{23.2} \\
 & GloFAS & 0.041 & 0.264 & 21.2 & +4.3 & 53.2 & 18.3 \\
 & persistence & -0.108 & 0.153 & 23.7 & +5.5 & 39.3 & 10.1 \\
\bottomrule
\end{tabular}
\end{table}

\subsection{Four-Variable and Discharge-Only Model Comparison}
\label{sec:ablation}

We compare the available four-variable and discharge-only models at 24-h lead to characterize their discharge performance.
The discharge-only model treats the other three land-surface variables as prescribed inputs and uses weighted MAE.
Because the models also differ in training schedule, input-data revision, and loss formulation, the comparison does not attribute the performance difference to multivariable prediction alone.

The four-variable model has lower MAE and RMSE and higher F1-score for floods at six of seven return periods in Table~\ref{tab:ablation}.
Discharge-only MAE is 32\% higher, at 2.07 versus 1.57~m$^3$\,s$^{-1}$, and RMSE is 27\% higher, at 41.14 versus 32.30~m$^3$\,s$^{-1}$.
The discharge-only model nevertheless has higher KGE and 100-yr F1, so the advantage depends on the metric.
The next comparison holds the four-variable architecture and 24-h training protocol fixed across three loss formulations.

% generated by paper/derive_ablation_multivar.py -- do not edit
%   Hapi (full)                              epoch 441 (run now at 610), evaluated 2026-08-30 00:35
%   -> Multivariable (discharge-only)        epoch 415 (run now at 420), evaluated 2026-09-01 00:05
\begin{table}[!htb]
\centering
\caption{Comparison of four-variable and discharge-only models at 24-h lead over CONUS in the 2024 test window. Training differences are described in Section~\ref{sec:ablation}. Full Hapi matches Table~\ref{tab:main_discharge}. MAE and RMSE are in m$^3$/s; other metrics are dimensionless. Bold marks the higher skill in each column.}
\label{tab:ablation}
\small
\setlength{\tabcolsep}{2.2pt}
\begin{tabular}{@{}lccccccccccc@{}}
\toprule
& & & & & \multicolumn{7}{c}{\textbf{F1 by GloFAS return period, yr}} \\
\cmidrule(lr){6-12}
\textbf{Variant} & MAE & RMSE & NSE & KGE & 1.5 & 2 & 5 & 10 & 20 & 50 & 100 \\
\midrule
\hapirow Full Hapi & \textbf{1.57} & \textbf{32.30} & \textbf{0.997} & 0.990 & \textbf{0.885} & \textbf{0.863} & \textbf{0.786} & \textbf{0.700} & \textbf{0.612} & \textbf{0.524} & 0.468 \\
Discharge-only & 2.07 & 41.14 & 0.996 & \textbf{0.992} & 0.869 & 0.842 & 0.761 & 0.672 & 0.593 & 0.522 & \textbf{0.481} \\
\bottomrule
\end{tabular}
\end{table}

\subsection{Loss-Weighting Ablation}
\label{sec:ablation_loss}

To distinguish the effect of task scaling from the effect of adapting the scales during training, we compare three loss formulations with the same four-variable architecture, inputs, targets, data splits, batch size, and 16-GPU training budget.
Uniform task weights minimize the unweighted mean of the four normalized-variable MAEs.
Fixed task scales use the Laplacian objective in Equation~\ref{eq:loss_mae}, with the four log-scales set to the values learned by the adaptive control and held constant throughout training.
Learned task scales use the same Laplacian objective but optimize the log-scales jointly with the model parameters.
Table~\ref{tab:ablation_loss} compares one trained model per loss formulation under the same single-step 24-h protocol on the 2024 test window.

Learned task scales yield the lowest discharge errors and highest F1 at every threshold.
Uniform weights increase MAE by 24\%, from 1.59 to 1.97~m$^3$\,s$^{-1}$, and reduce F1 by 4.7--11.5 percentage points, with the largest loss at the 100-yr threshold.
Fixed scales bring MAE to 1.64~m$^3$\,s$^{-1}$, only 3\% above the learned-scale result, accounting for most of the improvement in this metric.

Learning the task scales reduces MAE only slightly relative to fixed scales but produces larger gains in F1 at rare-event thresholds.
At the 100-yr threshold, fixed scales raise F1 from 0.345 to 0.382, whereas learning the scales raises it further to 0.460.
At every threshold from 10 to 100~yr, learning the scales adds more F1 than fixed scaling alone.

% generated by paper/derive_ablation_table.py -- do not edit
%   Learned adaptive task weighting (control)  epoch 953 (run now at 955), evaluated 2026-08-30 13:20
%   -> fixed prior weights                   epoch 953 (run now at 953), evaluated 2026-08-31 13:16
%   -> fixed precision                       epoch 952 (run now at 953), evaluated 2026-09-01 00:06
\begin{table}[!htb]
\centering
\caption{Loss-weighting ablation at 24-h lead over CONUS in the 2024 test window. The same four-variable architecture and data were used for single-step training on 16 GPUs, with checkpoints selected by minimum validation loss at epochs 952--953. Each loss-weighting variant is represented by one training run under these conditions. MAE and RMSE are in m$^3$/s; other metrics are dimensionless. These pretraining checkpoints precede the fine-tuning used in Table~\ref{tab:main_discharge}.}
\label{tab:ablation_loss}
\small
\setlength{\tabcolsep}{2.2pt}
\begin{tabular}{@{}lccccccccccc@{}}
\toprule
& & & & & \multicolumn{7}{c}{\textbf{F1 by GloFAS return period, yr}} \\
\cmidrule(lr){6-12}
\textbf{Variant} & MAE & RMSE & NSE & KGE & 1.5 & 2 & 5 & 10 & 20 & 50 & 100 \\
\midrule
Learned task scales & 1.59 & 30.94 & 0.998 & 0.990 & 0.877 & 0.856 & 0.779 & 0.690 & 0.601 & 0.513 & 0.460 \\
Uniform task weights & 1.97 & 38.71 & 0.996 & 0.985 & 0.830 & 0.802 & 0.710 & 0.607 & 0.505 & 0.410 & 0.345 \\
Fixed task scales & 1.64 & 31.77 & 0.997 & 0.987 & 0.856 & 0.835 & 0.748 & 0.647 & 0.544 & 0.445 & 0.382 \\
\bottomrule
\end{tabular}
\end{table}

\subsection{Non-Discharge Prognostic Skill}
\label{sec:prognostic_skill}

We evaluate the three non-discharge prognostic variables against persistence to determine whether they outperform carrying the initial state forward.
Against their reanalysis targets, Hapi predicts runoff and soil wetness with lower RMSE and MAE and higher NSE than persistence at all three leads in Table~\ref{tab:prognostic_skill}.
Runoff NSE remains positive, from 0.328 at 24~h to 0.298 at 72~h, whereas carrying the initial runoff field forward gives negative NSE at every lead.

The snow comparison depends on three cells with persistent target snow accumulation among the 349{,}402 evaluated cells.
The screening rule in Appendix~\ref{app:prognostic_protocol} excludes these cells, which contribute about 99\% of Hapi's unscreened squared error.
Retaining these three cells raises Hapi's 24-h RMSE from 2.51 to 22.20~mm and gives persistence lower RMSE and higher NSE at every lead.
Table~\ref{tab:prognostic_skill} reports both masks, showing the improvement across the screened snow field alongside the concentrated errors that reverse the full-mask ranking.

For soil wetness and screened snow, Hapi's advantage becomes clearer at longer leads.
Persistence is already accurate at 24~h for these slowly varying stores, but its RMSE increases more rapidly during the rollout.
By 72~h, Hapi reduces RMSE relative to persistence by 36\% for soil wetness and 32\% for screened snow.
Hapi also has lower MAE and higher NSE for both stores at every lead, while persistence retains higher 24-h KGE.

% Generated by paper/derive_prognostic_skill.py -- do not edit by hand.
\begin{table}[!htb]
\centering
\caption{Skill of the non-discharge prognostic variables against reanalysis over the 2024 test window. Hapi and analysis-time persistence use identical finite cell--day pairs on each variable's channel mask. Runoff and snow RMSE and MAE are in mm of water equivalent; soil wetness, NSE, and KGE are dimensionless. Bold marks the better score for each variable and lead. The screened snow rows exclude three persistent accumulation cells using the rule in Appendix~\ref{app:prognostic_protocol}.}
\label{tab:prognostic_skill}
\small
\setlength{\tabcolsep}{5pt}
\begin{tabular}{@{}lll
  S[table-format=2.4]
  S[table-format=1.4]
  S[table-format=-1.4]
  S[table-format=1.4]@{}}
\toprule
\textbf{Variable} & \textbf{Lead} & \textbf{Model}
  & {\textbf{RMSE}} & {\textbf{MAE}} & {\textbf{NSE}} & {\textbf{KGE}} \\
\midrule
\multicolumn{7}{@{}l}{\textit{Runoff water equivalent, 418{,}792 cells}} \\
\hapirow & 24\,h & Hapi & \textbf{1.8580} & \textbf{0.2850} & \textbf{0.3281} & 0.3474 \\
 &  & persistence & 2.4500 & 0.4363 & -0.1686 & \textbf{0.4156} \\
\hapirow & 48\,h & Hapi & \textbf{1.8900} & \textbf{0.2999} & \textbf{0.3107} & \textbf{0.3365} \\
 &  & persistence & 2.6790 & 0.5245 & -0.3862 & 0.3039 \\
\hapirow & 72\,h & Hapi & \textbf{1.9070} & \textbf{0.3105} & \textbf{0.2980} & \textbf{0.3327} \\
 &  & persistence & 2.7120 & 0.5615 & -0.4197 & 0.2871 \\
\addlinespace[3pt]
\multicolumn{7}{@{}l}{\textit{Snow water equivalent, 349{,}402 cells}} \\
\hapirow & 24\,h & Hapi & 22.2000 & 0.6048 & 0.9177 & 0.8711 \\
 &  & persistence & \textbf{2.7350} & \textbf{0.5889} & \textbf{0.9988} & \textbf{0.9964} \\
\hapirow & 48\,h & Hapi & 25.6900 & \textbf{0.8476} & 0.8897 & 0.8454 \\
 &  & persistence & \textbf{4.6310} & 1.0980 & \textbf{0.9964} & \textbf{0.9925} \\
\hapirow & 72\,h & Hapi & 28.1700 & \textbf{1.0500} & 0.8673 & 0.8260 \\
 &  & persistence & \textbf{6.2310} & 1.5660 & \textbf{0.9935} & \textbf{0.9883} \\
\addlinespace[3pt]
\multicolumn{7}{@{}l}{\textit{\quad screened, 349{,}399 cells}} \\
\hapirow & 24\,h & Hapi & \textbf{2.5100} & \textbf{0.5692} & \textbf{0.9977} & 0.9923 \\
 &  & persistence & 2.7350 & 0.5888 & 0.9972 & \textbf{0.9961} \\
\hapirow & 48\,h & Hapi & \textbf{3.4610} & \textbf{0.8040} & \textbf{0.9956} & \textbf{0.9947} \\
 &  & persistence & 4.6290 & 1.0980 & 0.9921 & 0.9915 \\
\hapirow & 72\,h & Hapi & \textbf{4.2460} & \textbf{1.0010} & \textbf{0.9933} & \textbf{0.9941} \\
 &  & persistence & 6.2290 & 1.5660 & 0.9856 & 0.9864 \\
\addlinespace[3pt]
\multicolumn{7}{@{}l}{\textit{Soil wetness index, 426{,}768 cells}} \\
\hapirow & 24\,h & Hapi & \textbf{0.0091} & \textbf{0.0037} & \textbf{0.9980} & 0.9965 \\
 &  & persistence & 0.0093 & 0.0042 & 0.9980 & \textbf{0.9986} \\
\hapirow & 48\,h & Hapi & \textbf{0.0104} & \textbf{0.0045} & \textbf{0.9974} & \textbf{0.9969} \\
 &  & persistence & 0.0147 & 0.0074 & 0.9949 & 0.9968 \\
\hapirow & 72\,h & Hapi & \textbf{0.0122} & \textbf{0.0057} & \textbf{0.9965} & \textbf{0.9957} \\
 &  & persistence & 0.0189 & 0.0101 & 0.9915 & 0.9950 \\
\bottomrule
\end{tabular}
\end{table}

\subsection{Per-Subset Skill}
\label{sec:results_subset}

For discharge, the continental F1 comparison combines rivers of different sizes and flood activity, leaving open whether Hapi's advantage holds across the river network.
We therefore compare Hapi and RiverMamba at their nominal forecast leads on AIFAS river points, flood-active cells, and three drainage-area groups separated at 500 and 5{,}000~km$^2$.
The first two subsets overlap the drainage groups; their definitions and cell counts are given in Appendix~\ref{app:splits}.
RiverMamba's latest discharge input is one day older at each nominal lead, as described in Section~\ref{sec:baselines}.

Table~\ref{tab:splits_f1} reports higher F1 for Hapi in all 75 comparisons across five subsets, three leads, and the five thresholds from 1.5 to 20~yr.
At the 2-yr threshold, the largest gaps occur in the small-river subset, where Hapi leads by 0.185--0.212 across the three leads.
When the 50- and 100-yr thresholds are added, four exceptions appear, all on large rivers at 48 and 72~h.
The 50-yr comparison at 48~h is effectively tied, at 0.2672 for RiverMamba against 0.2668 for Hapi.
At 72~h, RiverMamba's F1 is 0.187 versus Hapi's 0.090 at 50~yr and 0.131 versus 0.013 at 100~yr.
The broad advantage therefore weakens for rare exceedances on large rivers at longer leads.

Continuous discharge metrics test whether these spatial differences extend beyond threshold crossings.
Hapi has lower MAE on every subset at 24~h and on three of the five subsets at every lead, as shown in Appendix~\ref{app:splits}.
On large rivers, RiverMamba has lower MAE from 48~h and higher NSE at 72~h.
Its KGE is already higher at 24~h.

% Generated by derive_subset_f1_table.py from cached metrics and contingency counts.
\begin{table}[t]
\centering
\caption{Per-subset F1-score for floods at all seven primary return periods over CONUS in the 2024 test window. Hapi and RiverMamba, abbreviated RM, use their nominal forecast leads. RiverMamba's latest discharge input is one day older. Bold marks the higher value using unrounded scores.}
\label{tab:splits_f1}
\small
\setlength{\tabcolsep}{3.3pt}
\begin{tabular}{@{}llrccccccc@{}}
\toprule
\textbf{Subset} & \textbf{Lead} & \textbf{Model} & \textbf{1.5-yr} & \textbf{2-yr} & \textbf{5-yr} & \textbf{10-yr} & \textbf{20-yr} & \textbf{50-yr} & \textbf{100-yr} \\
\midrule
\hapirow AIFAS & 24\,h & Hapi & \textbf{0.895} & \textbf{0.870} & \textbf{0.785} & \textbf{0.690} & \textbf{0.585} & \textbf{0.461} & \textbf{0.366} \\
 &  & RM & 0.755 & 0.681 & 0.505 & 0.366 & 0.282 & 0.212 & 0.159 \\
\hapirow  & 48\,h & Hapi & \textbf{0.846} & \textbf{0.807} & \textbf{0.687} & \textbf{0.551} & \textbf{0.420} & \textbf{0.259} & \textbf{0.174} \\
 &  & RM & 0.722 & 0.641 & 0.459 & 0.321 & 0.236 & 0.158 & 0.101 \\
\hapirow  & 72\,h & Hapi & \textbf{0.816} & \textbf{0.767} & \textbf{0.633} & \textbf{0.481} & \textbf{0.344} & \textbf{0.182} & \textbf{0.116} \\
 &  & RM & 0.679 & 0.604 & 0.420 & 0.289 & 0.212 & 0.131 & 0.084 \\
\midrule
\hapirow Flood-active & 24\,h & Hapi & \textbf{0.917} & \textbf{0.878} & \textbf{0.803} & \textbf{0.726} & \textbf{0.647} & \textbf{0.570} & \textbf{0.523} \\
 &  & RM & 0.813 & 0.734 & 0.548 & 0.416 & 0.325 & 0.249 & 0.203 \\
\hapirow  & 48\,h & Hapi & \textbf{0.878} & \textbf{0.822} & \textbf{0.717} & \textbf{0.604} & \textbf{0.504} & \textbf{0.390} & \textbf{0.325} \\
 &  & RM & 0.784 & 0.696 & 0.506 & 0.375 & 0.283 & 0.201 & 0.152 \\
\hapirow  & 72\,h & Hapi & \textbf{0.851} & \textbf{0.787} & \textbf{0.666} & \textbf{0.539} & \textbf{0.436} & \textbf{0.311} & \textbf{0.258} \\
 &  & RM & 0.754 & 0.662 & 0.469 & 0.337 & 0.252 & 0.172 & 0.130 \\
\midrule
\hapirow Small rivers & 24\,h & Hapi & \textbf{0.877} & \textbf{0.854} & \textbf{0.775} & \textbf{0.683} & \textbf{0.590} & \textbf{0.504} & \textbf{0.454} \\
 &  & RM & 0.712 & 0.643 & 0.457 & 0.328 & 0.240 & 0.170 & 0.133 \\
\hapirow  & 48\,h & Hapi & \textbf{0.826} & \textbf{0.793} & \textbf{0.683} & \textbf{0.560} & \textbf{0.449} & \textbf{0.332} & \textbf{0.266} \\
 &  & RM & 0.684 & 0.609 & 0.416 & 0.289 & 0.199 & 0.127 & 0.091 \\
\hapirow  & 72\,h & Hapi & \textbf{0.796} & \textbf{0.758} & \textbf{0.633} & \textbf{0.500} & \textbf{0.388} & \textbf{0.269} & \textbf{0.216} \\
 &  & RM & 0.568 & 0.550 & 0.365 & 0.242 & 0.160 & 0.095 & 0.065 \\
\midrule
\hapirow Medium rivers & 24\,h & Hapi & \textbf{0.919} & \textbf{0.902} & \textbf{0.831} & \textbf{0.776} & \textbf{0.703} & \textbf{0.633} & \textbf{0.576} \\
 &  & RM & 0.781 & 0.727 & 0.570 & 0.419 & 0.333 & 0.262 & 0.228 \\
\hapirow  & 48\,h & Hapi & \textbf{0.869} & \textbf{0.838} & \textbf{0.734} & \textbf{0.625} & \textbf{0.524} & \textbf{0.400} & \textbf{0.334} \\
 &  & RM & 0.741 & 0.677 & 0.501 & 0.343 & 0.257 & 0.188 & 0.142 \\
\hapirow  & 72\,h & Hapi & \textbf{0.836} & \textbf{0.794} & \textbf{0.670} & \textbf{0.534} & \textbf{0.428} & \textbf{0.279} & \textbf{0.196} \\
 &  & RM & 0.706 & 0.636 & 0.447 & 0.285 & 0.224 & 0.160 & 0.118 \\
\midrule
\hapirow Large rivers & 24\,h & Hapi & \textbf{0.915} & \textbf{0.891} & \textbf{0.856} & \textbf{0.791} & \textbf{0.732} & \textbf{0.602} & \textbf{0.455} \\
 &  & RM & 0.848 & 0.781 & 0.699 & 0.586 & 0.536 & 0.441 & 0.273 \\
\hapirow  & 48\,h & Hapi & \textbf{0.840} & \textbf{0.792} & \textbf{0.725} & \textbf{0.595} & \textbf{0.507} & 0.2668 & 0.112 \\
 &  & RM & 0.791 & 0.707 & 0.590 & 0.445 & 0.374 & \textbf{0.2672} & \textbf{0.165} \\
\hapirow  & 72\,h & Hapi & \textbf{0.785} & \textbf{0.717} & \textbf{0.636} & \textbf{0.463} & \textbf{0.355} & 0.090 & 0.013 \\
 &  & RM & 0.748 & 0.666 & 0.537 & 0.398 & 0.301 & \textbf{0.187} & \textbf{0.131} \\
\bottomrule
\end{tabular}
\end{table}

\subsection{Forcing-Perturbation Sensitivity}
\label{sec:forcing_noise}

Operational forecasts depend on uncertain weather predictions rather than ERA5-Land forcing.
To test this sensitivity, we perturb precipitation at rollout steps 2 and 3 using four noise amplitudes ($\sigma=0$, 0.2, 0.5, and 1.0).
Appendix~\ref{app:forcing_perturbation} describes the perturbation construction and its effect on rainfall amounts.

F1 and NSE decrease or remain unchanged across the four tested amplitudes, with larger relative F1 losses at rare thresholds.
At 48~h and $\sigma=1$, NSE changes from 0.990 to 0.987, while 100-yr F1 falls from 0.269 to 0.039.
At the intermediate amplitude $\sigma=0.5$, the 100-yr F1 loss is already 38\%, compared with 4\% at the 1.5-yr threshold.
The small change in pooled NSE therefore conceals substantial sensitivity in forecasting floods at longer return periods.
Because these perturbations are not calibrated to a weather forecast system, operational skill still requires evaluation with archived NWP forecasts.

\begin{table}[!htb]
\centering
\caption{Sensitivity to precipitation perturbations at rollout steps 2--3 over CONUS in the 2024 test window.
The zero-preserving multiplier $e^{\sigma\varepsilon}$ uses a standardized Gaussian field smoothed with a spatial kernel of width 3 grid cells and a first-order autoregressive temporal coefficient of 0.8.
The log-noise amplitude $\sigma$ is distinct from the smoothing width.
The multiplier is not mean-centered; its nominal mean is $e^{\sigma^2/2}$.
The 24-h forcing is unchanged, and $\sigma=0$ is the unperturbed baseline.}
\label{tab:forcing_noise}
\small
\setlength{\tabcolsep}{6pt}
\begin{tabular}{@{}llcccc@{}}
\toprule
& & \multicolumn{4}{c}{\textbf{Log-noise amplitude $\sigma$}} \\
\cmidrule(lr){3-6}
\textbf{Lead} & \textbf{Metric} & 0.00 & 0.20 & 0.50 & 1.00 \\
\midrule
\multirow{5}{*}{48\,h} & F1 @ 1.5-yr & 0.833 & 0.827 & 0.800 & 0.709 \\
                       & F1 @ 5-yr   & 0.691 & 0.677 & 0.602 & 0.357 \\
                       & F1 @ 20-yr  & 0.460 & 0.442 & 0.334 & 0.114 \\
                       & F1 @ 100-yr & 0.269 & 0.254 & 0.168 & 0.039 \\
                       & NSE         & 0.990 & 0.990 & 0.990 & 0.987 \\
\midrule
\multirow{5}{*}{72\,h} & F1 @ 1.5-yr & 0.800 & 0.789 & 0.743 & 0.609 \\
                       & F1 @ 5-yr   & 0.638 & 0.616 & 0.507 & 0.244 \\
                       & F1 @ 20-yr  & 0.391 & 0.367 & 0.244 & 0.066 \\
                       & F1 @ 100-yr & 0.208 & 0.193 & 0.110 & 0.021 \\
                       & NSE         & 0.980 & 0.980 & 0.978 & 0.969 \\
\bottomrule
\end{tabular}
\end{table}

\subsection{Case Study: Hurricane Helene, 27 September 2024}
\label{sec:helene}

\begin{figure}[!htb]
\centering
\includegraphics[width=\textwidth]{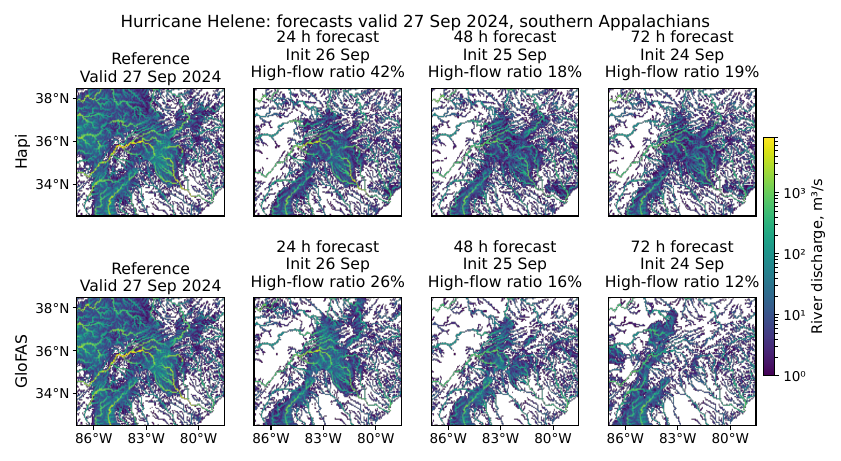}
\caption{Hurricane Helene discharge forecasts valid on 27 September 2024 over the southern Appalachians, spanning 32.5--38.5$^\circ$N and 87.0--78.5$^\circ$W, on a common logarithmic scale.
The top row shows Hapi and the bottom row GloFAS forecasts, each beside the GloFAS reanalysis reference.
The 24-, 48-, and 72-h forecasts were initialized on 26, 25, and 24 September, respectively.
The high-flow median ratio is the forecast median divided by the GloFAS reanalysis median over the same 204 cells in the highest 1\% of reference discharge; a value of 100\% would indicate equal medians.
The reference median is 3{,}547~m$^3$\,s$^{-1}$.}
\label{fig:helene}
\end{figure}

Hurricane Helene produced widespread high discharge across the southern Appalachians during the held-out test period.
Figure~\ref{fig:helene} compares Hapi and GloFAS forecasts valid on 27 September 2024 against GloFAS reanalysis using mapped discharge fields and the high-flow median ratio.
The 24-, 48-, and 72-h forecasts were initialized on 26, 25, and 24 September, respectively.
We quantify magnitude using the high-flow median ratio: the median forecast discharge over the 204 cells in the highest 1\% of reference discharge divided by the reanalysis median over those same cells.

The spatial median in GloFAS reanalysis over these cells is 3{,}547~m$^3$\,s$^{-1}$.
Hapi's high-flow median ratio is 42\%, 18\%, and 19\% at 24, 48, and 72~h, compared with 26\%, 16\%, and 12\% for GloFAS.
A ratio of 42\%, for example, means that the forecast median over the selected cells is 42\% of the reanalysis median.
Absolute magnitudes remain below half of the reference median for both systems.
During Helene, Hapi reproduces a larger fraction of the reference high-flow median at every lead, with the largest difference at 24~h.

\subsection{Computational Performance}
\label{sec:efficiency}

Hapi completes a three-step, four-variable 72-h forecast over the $512 \times 1152$ CONUS grid with an average inference time of 0.11\,s and 1.69\,GB of peak allocated memory, as reported in Table~\ref{tab:efficiency}.
RiverMamba produces seven discharge leads over its native 6.22-million-point global domain in 13.75\,s, with 13.33\,GB of peak allocated memory before CONUS extraction.
Both measurements exclude data loading, host-to-device transfer, output extraction, metric calculation, and file writing.
Training the final Hapi model through pretraining and multi-step fine-tuning used approximately 830 A100-hours.

Normalizing execution time by spatial units and output leads gives $0.062\,\mu$s per grid cell per lead for Hapi and $0.316\,\mu$s per river point per lead for RiverMamba.
Because the systems differ in domain, spatial representation, forecast outputs, GPU memory capacity, and software environment, these values measure implementation throughput rather than performance on equivalent workloads, as detailed in Appendix~\ref{app:implementation}.

Hapi controls computational scaling through fixed $8 \times 8$ attention windows, which keep attention cost linear in the number of spatial tokens, and hierarchical downsampling, which places the widest layers on smaller grids.
The resulting arithmetic estimate is 0.815 trillion multiply--accumulates per 24-h pass and 2.45 trillion for the three-step forecast; Appendix~\ref{app:blockcost} provides the derivation.
This estimate characterizes Hapi's computational structure and is distinct from the measured execution times above.

% Generated by paper/derive_efficiency_table.py -- do not edit by hand.
% Hapi job 54964.gpumgt1; RiverMamba job 55063.gpumgt1.
\begin{table}[!htb]
\centering
\caption{Average GPU-resident inference time over 10 repetitions at a batch size of 1 using 32-bit floating-point precision on one NVIDIA A100-SXM4 GPU, with 80\,GB capacity for Hapi and 40\,GB for RiverMamba. Normalized time is execution time divided by the number of spatial units and output leads. Memory is peak allocated memory. Data loading, transfer, CONUS extraction, metrics, and file writing are excluded.}
\label{tab:efficiency}
\small
\setlength{\tabcolsep}{2pt}
\begin{tabular}{@{}llrrrrr@{}}
\toprule
\textbf{System} & \textbf{Spatial units} & \textbf{Params} & \textbf{Leads} & \textbf{Time} & \textbf{Memory} & \textbf{Normalized time} \\
 & & M & & & GB & $\mu$s/unit/lead \\
\midrule
\hapirow Hapi & 589{,}824 grid cells & 43.96 & 3 & 109.5\,ms & 1.69 & 0.062 \\
RiverMamba & 6{,}221{,}926 river points & 4.38 & 7 & 13.75\,s & 13.33 & 0.316 \\
\bottomrule
\end{tabular}
\end{table}

\FloatBarrier

\section{Discussion}
\label{sec:discussion}

Hapi provides spatially distributed, multivariable hydrological forecasts over CONUS at 24--72-h lead times and maintains efficient continental-scale inference.
Its main advantage is flood forecasting, especially beyond the first forecast day.
Its ranking on continuous metrics, by contrast, depends on the scale of aggregation.
Together, the results show that learned task weighting, spatially distributed prediction, and efficient inference can be combined in a single continental model without limiting the forecast to discharge alone.

\paragraph{Continuous discharge versus flood forecasting skill.}
Continuous discharge metrics and the F1-score for floods rank the models differently because they emphasize different parts of the discharge distribution.
Pooled RMSE, NSE, and KGE give greater influence to large rivers, whereas median per-cell metrics describe performance at a typical location.
Hapi's median per-cell bias ratio in Appendix~\ref{app:percell} remains slightly below unity, indicating slight underprediction at a typical cell, whereas pooling the domain produces a positive bias.
The F1-score for floods measures whether forecasts and the reference agree on locally defined flood thresholds and is therefore more closely aligned with the warning decision motivating this study.
During Helene, Hapi has a higher high-flow median ratio than GloFAS at every lead, while both remain below 50\% of the reference median.

Hapi's F1 margin generally increases with lead time as persistence weakens.
Hapi has the lowest false-alarm ratio at the illustrated common and rare thresholds at every lead, although false alarms increase with lead for both Hapi and RiverMamba.
The advantage spans most of the river network but not the two rarest thresholds on large rivers, where RiverMamba is ahead in four comparisons at 48 and 72~h.
Absolute skill is low for both models in those comparisons, with no F1 above 0.27 in Table~\ref{tab:splits_f1}, and one of the four is the near-tie at the 50-yr threshold at 48~h.
Despite these exceptions, Hapi leads most threshold comparisons.

\paragraph{Spatial structure of continuous skill.}
The per-cell and independent gauge evaluations agree, and Hapi has higher NSE than GloFAS at about two-thirds of gauges.
This agreement shows that Hapi's advantage extends from the gridded reanalysis target to observed discharge at monitored rivers.
Performance nevertheless varies substantially among locations, with weaker relative skill clustered in the upper Midwest, Northeast, and mountain West, as shown in Figure~\ref{fig:percell_diff}.
Reservoir regulation is one plausible contributor because Hapi receives no storage state, operating rule, or release schedule, whereas LISFLOOD includes a reservoir module.
Although station names do not identify all regulated rivers or account for other basin differences, the observed association motivates further study of reservoir regulation as a source of spatial variation in forecast skill.

\paragraph{Learned weighting and multivariable prediction.}
The four-variable model performs better on most reported discharge metrics while producing all four fields, demonstrating the effectiveness of joint output prediction at continental scale.

In the loss comparison, adaptive task balancing gives the lowest discharge errors and highest F1 across the three tested loss formulations.
Fixed scales largely correct differences in target magnitude, whereas adapting the scales contributes more strongly at rare-flood thresholds.

For the three additional prognostics, Hapi outperforms persistence for runoff and soil wetness at every lead, while the snow result depends on the evaluation mask.
Three cells with persistent snow accumulation dominate Hapi's squared error and reverse the RMSE and NSE ranking on the full mask.
On the screened snow mask, Hapi also outperforms persistence throughout the rollout.

\paragraph{Sensitivity to meteorological forcing.}
Operational hydrological forecasts depend on uncertain weather predictions, whereas the main evaluation supplies ERA5-Land forcing over the forecast horizon.
Under the precipitation perturbations, pooled NSE changes little, but relative F1 losses grow at rarer thresholds.
Because the multiplier changes both rainfall variability and mean amount and is not calibrated to a forecast system, the experiment diagnoses sensitivity rather than predicting operational degradation.
Forecasting floods at longer return periods can therefore degrade before meteorological errors dominate aggregate continuous metrics.

\paragraph{Architecture and computational efficiency.}
Shifted-window attention and hierarchical downsampling expand the spatial receptive field while keeping attention local, and the mirrored decoder restores full-resolution fields without global attention.
Hapi's methodological contribution is an integrated system for continental multivariable hydrology that addresses fine output, continental extent, and heterogeneous prognostic targets.
Training is computationally intensive, but the low inference cost supports repeated continental forecasts once the model is trained.

\paragraph{Limitations and future directions.}
The evaluation spans a geographically diverse continental domain, but covers one 2024 test window and uses ERA5-Land fields over the forecast horizon.
Multi-year evaluation with archived operational NWP forecasts would test interannual and geographic robustness under realistic meteorological uncertainty.
The deterministic model can be extended with calibrated ensembles or a probabilistic prediction head to characterize forecast uncertainty.
Future analysis across basin characteristics and flow regimes could identify where additional hydrological inputs would provide the largest gains.

\paragraph{Operational relevance.}
Hapi combines continental coverage, multivariable prediction, strong flood forecasting performance, and low inference cost in one forecasting framework.
The forcing experiment shows that the F1-score for floods at longer return periods is particularly sensitive to meteorological input errors.
The independent USGS gauge evaluation shows that Hapi's median-NSE advantage extends to observed discharge at monitored rivers.
Its low execution cost makes repeated forecasts with updated meteorological inputs computationally practical and could support future ensemble prediction.

\section{Conclusion}
\label{sec:conclusion}

Hapi, a hierarchical Swin Transformer V2--based U-Net, forecasts discharge and evolving land-surface states together across CONUS at $0.05^{\circ}$ resolution.
The encoder--decoder combines fine three-dimensional patches, shifted-window attention, and multiscale skip connections, and learned task weights balance discharge, runoff, snow water equivalent, and soil wetness over 24--72-h forecasts.
In the 2024 CONUS potential-skill evaluation, Hapi leads nearly all F1 comparisons across the four models, and its advantage over RiverMamba remains significant after multiplicity correction throughout.
It also leads most SEDI comparisons with GloFAS and persistence, including every threshold at 48 and 72~h.
Although GloFAS leads several pooled metrics at longer leads, Hapi has the highest median per-cell NSE at every lead.
Independent validation at 3{,}881 USGS gauges confirms the same local-skill advantage against observations, with positive Hapi--GloFAS differences in median NSE under a regional-cluster bootstrap.

In the matched 24-h loss comparison, learned task weighting yields the lowest discharge errors and highest F1, with the largest F1 differences at rare thresholds.
Beyond discharge, Hapi outperforms persistence in RMSE, MAE, and NSE for runoff, soil wetness, and screened snow throughout the rollout.
The sensitivity of the F1-score for floods to precipitation perturbations makes forcing quality an important consideration for operational application.
Hapi's low inference cost makes future evaluation with rapidly updated meteorological inputs and ensemble forecasts computationally practical.
%The evaluation NetCDF, threshold-preparation pipeline, and trained model are intended for public archiving; the persistent repository identifier and release details must be added before publication.

% =============================================================
% Journal end-sections required by IOP author guidelines.
% \ack / \funding / \roles are user-supplied; \data is drafted here
% from the code + evaluation-artifact release.
% =============================================================

\ack{% TODO: acknowledgements text
We gratefully acknowledge the computing resources provided on Swing, a high-performance computing cluster operated by the Laboratory Computing Resource Center at Argonne National Laboratory.
}

\funding{% TODO: funder names and grant numbers
This work has been supported by the U.S. Department of Energy (DOE) Office of Cybersecurity, Energy Security, and Emergency Response (CESER) under DOE contract DE-AC02-06CH11357. Computing resources come from the Argonne Leadership Computing Facility, a DOE Office of Science user facility at Argonne National Laboratory.}

% \roles{% TODO: author contributions using CRediT taxonomy: https://credit.niso.org
% Author contribution statement to be provided by the authors using CRediT roles (e.g., Conceptualization, Methodology, Software, Formal analysis, Investigation, Data curation, Writing -- original draft, Writing -- review \& editing, Visualization, Supervision, Funding acquisition).
% }

\data{%
% The authors intend to archive the per-day predicted and target fields used to compute every table and figure in this paper as a single NetCDF file, \texttt{daily\_prognostics.nc}.
% It contains all four prognostic variables, river discharge, surface runoff, snow water equivalent, and the soil wetness index, as \texttt{<variable>\_pred} and \texttt{<variable>\_target} at each of the 24-, 48-, and 72-h leads, together with the persistence and GloFAS discharge baselines, all in physical units.
% The planned archive will also contain the model weights, the GloFAS return-level thresholds, the subset masks, and the evaluation and plotting scripts.
% Once archived, these files will permit reproduction of every reported number without rerunning inference or accessing the raw reanalysis inputs.
The source code and a versioned archive containing the evaluation artifacts, trained weights, and analysis code is being prepared for public release.
Reference ERA5-Land and GloFAS reanalysis inputs, needed for retraining or a fresh inference pass, are publicly available from the Copernicus Climate Change Service at \url{https://cds.climate.copernicus.eu} and the GloFAS portal at \url{https://global-flood.emergency.copernicus.eu}.
The upstream-drainage-area field is distributed with the GloFAS v4.0 auxiliary data~\citep{ecmwf2023glofasaux}.
Representative-gauge hydrographs are given in Figure~\ref{fig:hydrographs} and per-cell skill maps in Appendix~\ref{app:percell}; spatial flood-risk maps at the 100-yr return period are deferred to a future release.
}

\bibliographystyle{plainnat}
\bibliography{references}
\clearpage
\appendix

\begin{center}
{\Large\bfseries Appendix}
\end{center}

\titlelabel{Appendix~\thetitle\quad}

\section{Per-Variable Normalization}
\label{app:normalization}

\begin{figure}[h]
\centering
\includegraphics[width=\textwidth]{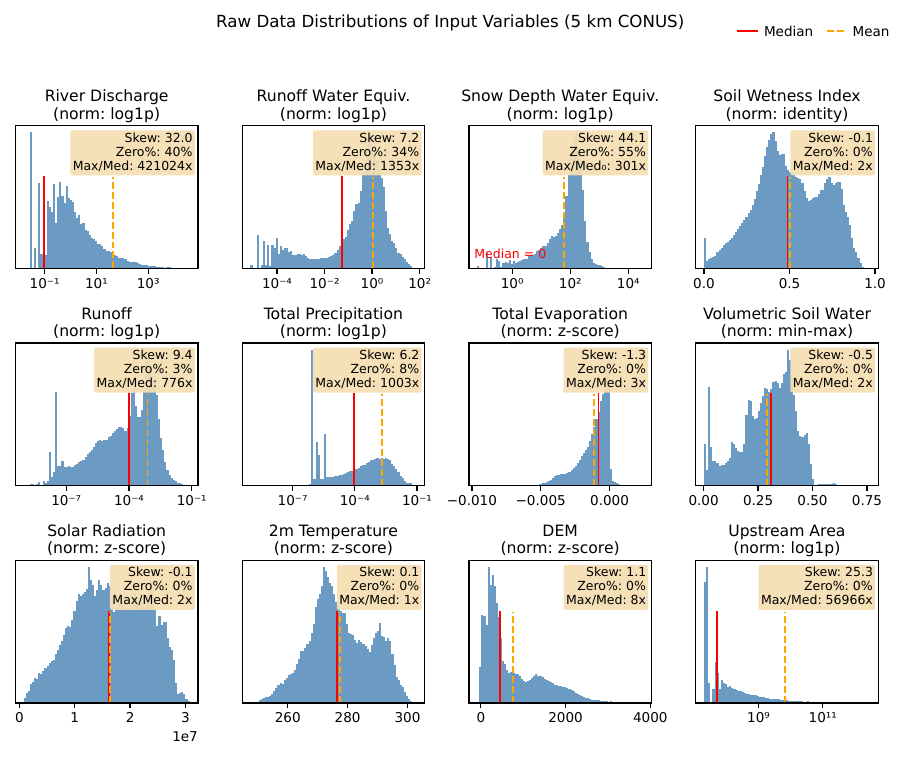}
\caption{Raw distributions of the 12 sampled input variables over the 419{,}682-cell 2\,m-temperature mask.
This illustration uses a shorter window than the descriptive statistics in Table~\ref{tab:normalization}.
Each panel shows skewness, zero fraction, and the maximum divided by the absolute median. Med$_0$ denotes the median of nonzero values when the overall median is zero.
Variables normalized with log1p use a logarithmic $x$-axis to show the tail structure.
Red and orange lines mark the median and mean. Large separation indicates high skewness.}
\label{fig:distributions}
\end{figure}

The 12 input variables have very different scales and distributions, so preprocessing is specified separately for each variable.
We use the following transformations for the distributions shown in Figure~\ref{fig:distributions}:

\begin{itemize}[nosep]
    \item \textbf{log1p}: $x \mapsto \log(1+x)$, without subsequent standardization.
    Used for variables with strong right skew and dynamic ranges spanning several orders of magnitude (e.g., river discharge has skewness $>30$ and a max/median ratio above $10^5$).
    \item \textbf{min-max}: $x \mapsto (x - x_{\min}) / (x_{\max} - x_{\min})$.
    Used for volumetric soil water.
    \item \textbf{z-score}: $x \mapsto (x - \mu) / \sigma$.
    Used for variables that are close to Gaussian (e.g., 2\,m temperature, solar radiation).
    \item \textbf{identity}: $x \mapsto x$.
    Used for the soil wetness index, which already lies in $[0,1]$.
\end{itemize}

\noindent The normalization choice and key statistics for each variable are provided in Table~\ref{tab:normalization}.
Total evaporation retains ERA5-Land's downward-positive flux convention and is standardized without a logarithmic transform.

\begin{table}[h]
\centering
\caption{Per-variable normalization choices and descriptive distributional statistics. The model's normalization parameters were estimated from the 2005--2022 training period only. The descriptive statistics were computed over the full 2005--2024 record (every tenth day) on the $0.05^{\circ}$ CONUS grid.
The statistics use the 419{,}682 cells where 2\,m temperature is defined. Flood forecasting uses the 414{,}795-cell discharge mask described in Appendix~\ref{app:masking}.
Skewness is the standard biased estimator. Zero\,\% counts exact zeros, and Max/Med is the maximum divided by the absolute median.
$^{\dagger}$Snow depth has a zero median, so its Max/Med ratio uses the median of nonzero values.}
\label{tab:normalization}
\small
\setlength{\tabcolsep}{4pt}
\begin{tabular}{@{}llrrrl@{}}
\toprule
\textbf{Variable} & \textbf{Norm} & \textbf{Skew} & \textbf{Zero\,\%} & \textbf{Max/Med} & \textbf{Rationale} \\
\midrule
River discharge         & log1p  & 38.9  & 35\% & $7.4 \times 10^5$ & Extreme skew, 6 orders of magnitude \\
Runoff water equiv.     & log1p  & 10.5  & 33\% & $4.1 \times 10^3$ & Heavy right skew \\
Snow water equivalent   & log1p  & 118.5 & 78\% & $8.8 \times 10^2$$^{\dagger}$ & Extreme skew and seasonal absence \\
Soil wetness index      & identity & $-$0.0 & 0\%  & 2.2              & Already bounded in $[0,1]$ \\
Runoff                  & log1p  & 11.7  & 4\%  & $2.1 \times 10^3$ & Heavy right skew \\
Total precipitation     & log1p  & 5.8   & 8\%  & $3.1 \times 10^3$ & Heavy right skew \\
Total evaporation       & z-score & $-$1.0 & 0\%  & 3.9             & Signed flux, standardized directly \\
Volumetric soil water   & min-max & $-$0.2 & 0\%  & 2.8              & Bounded fraction, symmetric \\
Solar radiation         & z-score & $-$0.0 & 0\%  & 2.1              & Nearly symmetric \\
2\,m temperature        & z-score & $-$0.4 & 0\%  & 1.1              & Nearly Gaussian \\
DEM                     & z-score & 1.1   & 0\%  & 8.4              & Static terrain elevation \\
Upstream area           & log1p  & 25.3  & 0\%  & $5.7 \times 10^4$ & 5 orders of magnitude \\
\bottomrule
\end{tabular}
\end{table}

\section{Physical Masking}
\label{app:masking}

The CONUS grid at $0.05^{\circ}$ resolution contains ocean pixels, the Great Lakes, and zero-padded regions outside the domain boundary.
We compute a separate mask $\mathbf{m}_c \in \{0, 1\}^{H \times W}$ for each channel $c$:
\begin{equation}
    m_{c,h,w} = \mathbf{1}\bigl[\exists\, t \text{ s.t. } x_{c,t,h,w} \neq 0 \bigr].
\end{equation}
A pixel is included if it is nonzero at any time step in the 2005--2024 record.
Different variables produce different masks.
The masks used during training and rollout contain 414{,}795 discharge cells, 419{,}054 runoff cells, 398{,}288 snow cells, and 426{,}768 soil-wetness cells.
Table~\ref{tab:normalization} and Figure~\ref{fig:distributions} use the 419{,}682 cells with valid 2\,m temperature to summarize the input distributions.
Discharge evaluation uses the 414{,}795-cell discharge mask.
Static fields (DEM, upstream area) use a simpler rule: the mask is 1 wherever the field value is nonzero.

These masks serve two purposes:
\begin{enumerate}[nosep]
    \item \textbf{Loss masking}: masked-out pixels do not contribute to the training loss, so the model does not spend capacity on zero-padded regions.
    \item \textbf{Output zeroing}: predictions are multiplied by the mask before being fed back during autoregressive rollout, keeping zero-padded regions exactly at zero.
\end{enumerate}

\paragraph{Preparation of the flood thresholds.}
The GloFAS v4.0 return-level fields~\citep{ecmwf2023glofasaux} are distributed globally at $0.05^{\circ}$ on a $3000 \times 7200$ grid whose cell centers coincide with those of our $512 \times 1152$ CONUS grid.
Preparation therefore requires only coordinate subsetting: we select the CONUS cells by exact coordinate match.
We verified that the maximum coordinate offset between target and source is $0$, that the subset values are bit-identical to the source, and that the missing-data pattern is preserved exactly.
We avoid interpolation because it would smooth the return-level field, including the local maxima that define the scored events.
Return levels are defined at $426{,}768$ of the $589{,}824$ cells in the CONUS box, the same cells covered by the GloFAS soil-wetness field.
The 414{,}795-cell discharge mask lies within this domain and is used for discharge evaluation.
Cells left undefined in the source are assigned $+\infty$ at load time so that no event can ever be triggered on them.

\section{Physical Bounds on Prognostic Feedback}
\label{app:denoising}

Several prognostic variables (river discharge, runoff, and snow water equivalent) are physically nonnegative and carry large fractions of exact zeros. The soil wetness index is bounded in $[0,1]$ by construction.
During autoregressive rollout, predictions that stray outside those ranges are fed back as input and accumulate across steps.
Each prognostic channel therefore has optional physical bounds, and both bounds are enforced before the prediction is fed back.

The constants are computed in physical units and then mapped through the channel's own normalization $f_c$ from Appendix~\ref{app:normalization}, using $\log(1+x)$ for discharge and runoff.
Using the standardization formula $(x - \mu_c)/\sigma_c$ would place the thresholds incorrectly for these variables.

The floor uses snapping because the targets have real mass at the floor.
For a channel with declared minimum $l_c$ and smallest training value $z_c$ strictly above it, the cutoff is the midpoint between the two,
\begin{equation}
    \tau_c = f_c\!\left(\tfrac{1}{2}\left(l_c + z_c\right)\right),
\end{equation}
and any prediction below $\tau_c$ is replaced by $f_c(l_c)$.
For the nonnegative variables $l_c = 0$, so the cutoff in physical units is half the smallest nonzero value observed in training.
Predictions below this cutoff are set to zero.
For a channel with no observed value above its lower bound, the rule reduces to standard clipping.

The upper bound uses standard clipping: a prediction above a declared maximum $u_c$ is set to $f_c(u_c)$, while all other predictions remain unchanged.
Snapping up would be wrong wherever no target reaches the ceiling, as with the soil wetness index, which is bounded at 1 but attains 0.988 at most in the data.

Both rules are applied at intermediate rollout steps only, in training and at inference alike, so the two see the same feedback.
Neither is applied at the final step, where the loss is taken, so the loss sees raw output.
At intermediate steps the clip does zero the gradient on the cells it moves, which is a cost we accept in exchange for training and inference following the same rollout.

\section{Two-Phase Training}
\label{app:twophase}

Training uses the objective in Section~\ref{sec:loss} under the following two-phase schedule.

\paragraph{Phase 1: single-step pretraining.}
The model is trained at a single autoregressive step ($J = 1$) under Eq.~\ref{eq:loss_mae}, on a cosine learning-rate schedule with warm restarts.
This fixes the one-step operator before the compounding errors of rollout are introduced.

\paragraph{Phase 2: multi-step fine-tuning.}
The pretrained weights are then fine-tuned at randomized rollout depth $J \sim \mathrm{Uniform}\{1,\dots,J_{\max}\}$ with $J_{\max} = 3$, backpropagating through the whole rollout, as described in Section~\ref{sec:loss}.
Intermediate predictions are held to their declared physical bounds before feedback, as specified in Appendix~\ref{app:denoising}, so the model sees its own accumulated errors under the same rule used at inference.

Phase~2 uses two constraints.
Phase~2 uses a base learning rate of $1{\times}10^{-5}$, compared with $5{\times}10^{-4}$ in Phase~1, and keeps the learned log-scales $s_i$ fixed.
If left trainable, the log-scales can reduce the rollout objective by increasing $s_i$ for variables with accumulating multi-step error and incurring the corresponding regularization penalty.
In Phase~1, trainable log-scales adjust the relative contributions of the four variables to the objective.
During Phase~2, the same adjustment could lower the objective without improving rollout predictions, so the log-scales are held fixed.

\section{Architecture Hyperparameter Sweep}
\label{app:sweep}

We ran a Bayesian hyperparameter sweep over 224 hyperparameter combinations (206 completed) using Weights \& Biases, with a budget of 20 epochs per run.
The search space is specified in Table~\ref{tab:sweep_space}.

\begin{table}[h]
\centering
\caption{Hyperparameter search space. Depth schedules list the encoder stages only; the decoder mirrors the encoder around the shared bottleneck, which is counted once.}
\label{tab:sweep_space}
\small
\begin{tabular}{@{}ll@{}}
\toprule
\textbf{Hyperparameter} & \textbf{Values} \\
\midrule
Encoder depth schedule  & $(1,1,4)$, $(1,2,4)$, $(1,2,8)$, $(2,2,4)$, $(2,2,8)$, $(2,3,8)$, $(3,3,8)$ \\
Embedding dimension & 48, 96, 192, 240 \\
Patch size $p$  & 2, 4 \\
Window size $w$ & 4, 8 \\
\bottomrule
\end{tabular}
\end{table}

Two architectural regimes stood out.
\textbf{Shallow encoders with fine-grained tokenization} ($p{=}2$, depths $(1,1,4)$ or $(1,2,4)$, $d{=}192$--$240$) dominated the top 10 runs by validation loss and converged fastest.
\textbf{Deep encoders with coarser tokenization} ($p{=}4$, depths $(3,3,8)$, $d{=}192$--$240$) formed a competitive second tier that converged more slowly but used less compute per step.
Window size had little effect on final performance.
Embedding dimensions of 192 and 240 consistently beat 48 and 96, showing that model capacity matters at this spatial resolution.

\section{Implementation Details}
\label{app:implementation}

\paragraph{Optimization and training schedule.}
We train on 16 NVIDIA A100 GPUs distributed across two eight-GPU nodes.
We use AdamW with $\beta_1{=}0.9$, $\beta_2{=}0.999$, and two parameter groups.
Model parameters use base learning rate $5{\times}10^{-4}$ during single-step pretraining and $1{\times}10^{-5}$ during multi-step fine-tuning.
A 2-epoch linear warmup is used for pretraining. Cosine cycles span 20 epochs in pretraining and 500 epochs in multi-step fine-tuning, with weight decay $1{\times}10^{-8}$.
Pretraining uses batch size 6 per GPU. Multi-step fine-tuning uses batch size 4 per GPU and $J_{\max}{=}3$.
The learning rate is scaled by $\sqrt{B \times \text{world\_size}}$ (square-root scaling for AdamW).
No gradient clipping is applied.

\paragraph{Adaptive task weighting.}
The log-scale parameters $\{s_c\}$ in the adaptive task-weighting loss use a fixed learning rate of $10^{-3}$ with no weight decay, keeping them tied to per-variable error magnitudes throughout the model's annealing schedule.
Under multi-step fine-tuning the log-scales are frozen, requiring the model to reduce the multi-step rollout error directly.
Per-variable prior weights are uniform ($w_i {=} 1$) in every reported run. The objective supports non-uniform priors, but they are not used.
Initial log-scales are $s_i^{(0)} {=} (-4.24,\,-3.17,\,-3.00,\,-4.48)$ for (discharge, surface runoff, snow water equivalent, soil wetness index), seeded from per-variable error magnitudes observed in a short warmup run.
The evaluated checkpoint retains $s_i {=} (-4.4134,\,-3.7854,\,-4.4622,\,-6.2489)$, giving inverse scales $\exp(-s_i)$ of $(82.6,\,44.1,\,86.7,\,517.4)$.
These log-scales are inherited unchanged from the pretraining checkpoint.
The learned weighting is far from uniform and would be hard to guess.
Relative to discharge, the soil wetness index is weighted $6.3\times$ and surface runoff $0.53\times$.

\paragraph{Data loading.}
Each input variable is stored as a separate NumPy memory-mapped file with shape $(T_{\text{total}}, H, W)$.
DataLoaders use 2 workers with a prefetch factor of 1 and persistent workers to reduce startup cost.
Per-variable normalization statistics are computed once and cached to disk.

\paragraph{Distributed training.}
The training script auto-detects the distributed environment (SLURM, torchrun, or JSM) and falls back to single-GPU training if initialization fails.
Logging and checkpointing run only on the main process. Per-node file operations, such as memory-mapped file copies to local scratch, run once per node.

\paragraph{Hapi benchmark protocol.}
We benchmarked the model of record on one NVIDIA A100-SXM4-80GB with driver 580.95.05 using PyTorch 2.10.0+cu128.
All benchmark runs used FP32 arithmetic with PyTorch autocast disabled.
We enabled \texttt{torch.compile(mode="default")}, which optimizes the model's computation graph before repeated execution.
CUDA events measured the complete rollout, including forward evaluation, masking, physical bounds, prognostic feedback, and forcing injection.
Measurements began after data loading and host-to-device transfer and ended before metric computation and file writing.
Each forecast-length and batch-size combination used 3 warm-up rollouts and 10 timed repetitions.
Peak allocated memory from \texttt{torch.cuda.max\_memory\_allocated} includes the model and device-resident inputs.

\paragraph{Hapi benchmark results.}
The 43.96\,M-parameter model requires one forward pass for a 24-h forecast and three autoregressive passes for a 72-h forecast.
At batch~1, average inference time increases from 36.6\,ms at 24\,h to 109.5\,ms at 72\,h, with interquartile ranges of 36.5--36.6 and 109.4--109.5\,ms, respectively.
Peak allocated memory increases from 1.63 to 1.69\,GB.
At batch size 4, the corresponding average inference times are 139.2 and 418.5\,ms, with peak allocated memory of 4.49 and 4.74\,GB.
The interquartile range remained below 0.3\% of the average across all four forecast-length and batch-size combinations.

\paragraph{RiverMamba timing protocol.}
We measured the public RiverMamba checkpoint with 4.38\,M parameters in FP32 on one NVIDIA A100-SXM4-40GB using PyTorch 2.4.1+cu124.
As for Hapi, graph optimization was enabled before timing: \texttt{torch.compile(mode="default")} transforms supported operations into optimized GPU graphs, while RiverMamba's custom selective-scan and causal-convolution operators remain eager operations.
RiverMamba produces seven leads jointly for 6{,}221{,}926 global river points, processed in the same 20 chunks used for its native global predictions.
We summed CUDA-event execution times across all chunks for each complete global forecast.
After 3 warm-up forecasts, 10 repetitions gave an average of 13.75\,s, with an interquartile range of 13.748--13.749\,s, and peak allocated memory of 13.33\,GB.
Input preparation, host-to-device transfer, CONUS extraction, metric calculation, and file writing were excluded from both systems' measurements.

The two runs used the same A100-SXM4 compute architecture but different memory capacities and software versions.
Because Hapi predicts four variables on a regular CONUS grid while RiverMamba predicts discharge on global river points, the raw times do not represent equal workloads.
Table~\ref{tab:efficiency} therefore also reports execution time divided by spatial units and output leads.
This normalization measures implementation throughput. It does not remove differences in model structure, spatial representation, or forecast construction.

\section{Hapi Arithmetic Estimate}
\label{app:blockcost}

We count one multiplication and its accumulation as one multiply--accumulate (MAC). Counting them separately gives twice as many floating-point operations.
For $N$ tokens, channel width $C$, MLP expansion $r=4$, and window width $w=8$, a Hapi block requires
\begin{equation}
    \mathrm{MAC}_{\mathrm{block}} = 13NC^{2} + 2Nw^{2}C.
\end{equation}
The first term covers the query--key--value--gate, output, and MLP projections. The second covers the two attention matrix products.
We omit normalization, activation, softmax, rotary embedding, masking, and data movement.
Patch size 2 converts the $512 \times 1152$ input into 147{,}456 tokens. Two merging steps reduce this count to 36{,}864 and 9{,}216.
Table~\ref{tab:blockcost} gives the resulting component totals.
The complete encoder--decoder requires 814.9 billion MACs per 24-h pass and 2.45 trillion MACs for the three-pass forecast.
The two attention matrix products contribute 2.5\% of the estimated MACs within Hapi's transformer blocks. Their cost is linear in the number of spatial tokens because the window size is fixed.

\begin{table}[h]
\centering
\caption{Estimated Hapi arithmetic for one 24-h forward pass. The three backbone rows include both encoder and decoder blocks at each resolution.}
\label{tab:blockcost}
\small
\begin{tabular}{@{}lrrr@{}}
\toprule
\textbf{Component} & \textbf{Tokens} & \textbf{Width} & \textbf{MACs, billion} \\
\midrule
 Backbone, 2 blocks & 147{,}456 & 192 & 148.6 \\
 Backbone, 4 blocks & 36{,}864 & 384 & 289.9 \\
 Backbone, 4 blocks & 9{,}216 & 768 & 286.3 \\
 Patch embedding & 147{,}456 & 192 & 2.7 \\
 Merge, unmerge, and skip projections & --- & --- & 65.2 \\
 Output expansion and projection & --- & --- & 22.2 \\
\midrule
 \textbf{One 24-h pass} & & & \textbf{814.9} \\
 \textbf{Three-pass 72-h forecast} & & & \textbf{2{,}444.8} \\
\bottomrule
\end{tabular}
\end{table}

\section{Evaluation Metric Definitions}
\label{app:metric_definitions}

Let $y_i$ and $\hat{y}_i$ denote the reference and forecast values on the $N$ paired finite samples used in a calculation, and let an overbar denote the sample mean.
The continuous metrics are
\begin{align}
    \mathrm{MAE} &= \frac{1}{N}\sum_{i=1}^{N}\lvert \hat{y}_i-y_i\rvert, \\
    \mathrm{RMSE} &= \sqrt{\frac{1}{N}\sum_{i=1}^{N}(\hat{y}_i-y_i)^2}, \\
    \mathrm{NSE} &= 1-\frac{\sum_{i=1}^{N}(\hat{y}_i-y_i)^2}
    {\sum_{i=1}^{N}(y_i-\bar{y})^2}, \\
    \mathrm{KGE} &= 1-\sqrt{(r-1)^2+(\alpha-1)^2+(\beta-1)^2},
    \label{eq:evaluation_continuous}
\end{align}
where $r$ is the Pearson correlation between $\hat{y}$ and $y$, $\alpha=s_{\hat{y}}/s_y$ is the variability ratio, and $\beta=\bar{\hat{y}}/\bar{y}$ is the bias ratio~\citep{gupta2009kge}.
For the gauge evaluation, percent bias is
\begin{equation}
    \mathrm{PBIAS}=100\frac{\sum_{i=1}^{N}(\hat{y}_i-y_i)}{\sum_{i=1}^{N}y_i}.
\end{equation}
NSE is undefined when the reference variance is zero.
KGE is undefined when the reference mean or either standard deviation needed by its ratios and correlation is zero.
PBIAS is undefined when the summed reference discharge is zero.

For flood forecasting, a forecast and reference are each classified as floods when their discharge reaches or exceeds the same local return level.
The resulting contingency table contains true positives (TP), false positives (FP), false negatives (FN), and true negatives (TN).
We define
\begin{align}
    \mathrm{precision} &= \frac{\mathrm{TP}}{\mathrm{TP}+\mathrm{FP}}, &
    \mathrm{recall}=H &= \frac{\mathrm{TP}}{\mathrm{TP}+\mathrm{FN}}, \\
    \mathrm{F1} &= \frac{2\,\mathrm{TP}}{2\,\mathrm{TP}+\mathrm{FP}+\mathrm{FN}}, &
    \mathrm{CSI} &= \frac{\mathrm{TP}}{\mathrm{TP}+\mathrm{FP}+\mathrm{FN}}, \\
    \mathrm{FAR} &= \frac{\mathrm{FP}}{\mathrm{TP}+\mathrm{FP}}, &
    F &= \frac{\mathrm{FP}}{\mathrm{FP}+\mathrm{TN}}.
    \label{eq:evaluation_categorical}
\end{align}
Here, $H$ is the hit rate, also called the probability of detection (or recall), FAR is the false-alarm ratio, and $F$ is the false-alarm rate (also called the probability of false detection).
The frequency bias shown by the rays in Figure~\ref{fig:performance_diagram} is
\begin{equation}
    B=\frac{\mathrm{TP}+\mathrm{FP}}{\mathrm{TP}+\mathrm{FN}}
     =\frac{\mathrm{recall}}{\mathrm{precision}}.
\end{equation}
The symmetric extremal dependence index is
\begin{equation}
    \mathrm{SEDI}=
    \frac{\ln F-\ln H-\ln(1-F)+\ln(1-H)}
         {\ln F+\ln H+\ln(1-F)+\ln(1-H)}
    \label{eq:sedi}
\end{equation}
for $0<H<1$ and $0<F<1$~\citep{ferro2011extremal}.
We report SEDI as undefined when either rate is 0 or 1 rather than clipping the rate before evaluation.

Unless stated otherwise, the main discharge metrics pool all valid cell--day pairs over the CONUS mask before each score is formed.
Likewise, pooled flood forecasting scores sum TP, FP, FN, and TN over valid cells and days before applying the equations above.
The per-cell analysis in Appendix~\ref{app:percell} instead computes a score independently at each cell and summarizes the resulting spatial distribution.
The gauge analysis computes each score independently at each retained gauge and reports medians across gauges.

\section{Precipitation-Perturbation Protocol}
\label{app:forcing_perturbation}

At rollout steps 2 and 3, precipitation is multiplied by $e^{\sigma\varepsilon}$, where $\varepsilon$ is a standardized Gaussian field correlated in space and across rollout steps.
This intervention preserves dry cells and scales the perturbation with rainfall intensity.
We test $\sigma=0$, 0.2, 0.5, and 1.0.
The 24-h forcing is unchanged, and the zero-amplitude run recovers the unperturbed baseline.

The log-normal multiplier changes both rainfall variability and mean amount.
Its median is one, but its mean is $e^{\sigma^2/2}$, reaching about 1.65 at $\sigma=1$.
Within one standard deviation of zero, the perturbations correspond to changes of $-18\%$ to $+22\%$ at $\sigma=0.2$ and $-39\%$ to $+65\%$ at $\sigma=0.5$.
The resulting percentage changes are asymmetric about zero.

\section{Gauge Matching and Hydrograph Selection}
\label{app:gauge_protocol}

We match each USGS gauge to the cell within $\pm2$ grid cells whose GloFAS v4.0 upstream drainage area best matches the reported gauge basin area.
This search reduces mismatches between adjacent tributaries and main-stem reaches that coordinate-only matching can introduce.
Matches differing by more than a factor of two in drainage area are rejected, leaving 7{,}838 of 7{,}881 candidates with a median area ratio of 1.01.
Where multiple gauges claim one cell, we retain the closest drainage-area match before checking observation availability, leaving 7{,}256 unique cells.
Observations are available for 3{,}976 retained gauges. Requiring at least 200 paired days leaves 3{,}881 scored gauges.
Paired finite NSE values are available at 3{,}869 gauges at each lead.

We quantify uncertainty in the model-specific median NSE differences with a paired cluster bootstrap over two-digit hydrologic regions (HUC2), rather than treating gauges as spatially independent.
Of the 3{,}869 paired gauges, 3{,}863 have HUC2 labels spanning 18 regions; six international-boundary gauges without HUC labels are excluded only from this uncertainty analysis.
For each of 10{,}000 replicates, we sample 18 HUC2 regions with replacement, retain all paired model scores within each sampled region, and recompute the difference between the two model-specific gauge medians.
The 2.5th and 97.5th percentiles form the 95\% interval; the random seed is 20260927.
For Hapi minus GloFAS, the median-NSE differences (95\% intervals) are 0.085 (0.046--0.105), 0.101 (0.046--0.128), and 0.120 (0.047--0.172) at 24, 48, and 72~h.
For Hapi minus persistence, they are 0.091 (0.045--0.110), 0.196 (0.107--0.292), and 0.269 (0.158--0.386).
These intervals quantify regional sampling variability within the CONUS test domain.

Hydrograph selection uses Hapi's 72-h NSE and a stricter requirement of at least 300 paired days, leaving 3{,}710 gauges with finite NSE before regional and drainage-area selection.
We take the three most populated eligible water-resource regions and choose the gauge nearest each region--area bin's median NSE.
The drainage-area bands are 500--5{,}000, 5{,}000--50{,}000, and at least 50{,}000~km$^2$.
Eligible bin counts from smallest to largest are 267, 149, and 60 in Missouri, 304, 117, and 1 in South Atlantic--Gulf, and 247, 89, and 27 in the Pacific Northwest.
The selected gauge approximates a bin's median skill. The single-gauge Alabama River bin cannot represent within-bin variability.

\section{Non-Discharge Evaluation Masks and Snow Screening}
\label{app:prognostic_protocol}

Each variable is evaluated on its own prediction mask, reconstructed from cells with nonzero temporal variance in the 24-h prediction.
Masked outputs remain constant at the inverse transform of zero, which is zero for the three non-discharge prognostic variables.
Hapi and analysis-time persistence are scored on identical finite cell--day pairs at each lead.
The original GloFAS GRIB headers specify kg\,m$^{-2}$ for both runoff and snow water equivalent. Preprocessing retains these numerical values, which are equivalent to millimetres of water equivalent. Runoff and snow errors are therefore reported in millimetres without numerical rescaling. Soil wetness is dimensionless.

The snow sensitivity analysis excludes cells whose minimum finite target snow water equivalent exceeds 100~mm over the available evaluation window.
This target-based rule removes three of 349{,}402 snow cells and does not use prediction error to select them.
The scoring script applies the rule to the evaluation target, so the screened scores are a sensitivity analysis conditional on that mask.
The excluded cells retain large snow accumulations in the Cascades and account for about 99\% of unscreened squared error.
They remain in training and in the full-mask evaluation. Table~\ref{tab:prognostic_skill} reports both masks.

\section{Extended-Lead RiverMamba Diagnostics}
\label{app:rm_extended}

Hapi is trained and evaluated at lead times of 24, 48, and 72 hours, so the head-to-head comparisons in Section~\ref{sec:experiments} stop at 72 hours.
Table~\ref{tab:rm_extended} reports RiverMamba's F1-scores for floods at 4--7-day leads on the same CONUS test window and return periods.
The numbers come from the public RiverMamba checkpoint evaluated under our pipeline. They are \emph{not} a head-to-head comparison with Hapi at those leads.

F1 generally declines as lead time increases beyond the leads in Table~\ref{tab:flood_f1}.
At the 1.5-yr threshold the CONUS-pooled series runs $0.728 \to 0.697 \to 0.592$ over the 24-, 48-, and 72-h leads of the main text, then $0.580 \to 0.566 \to 0.571 \to 0.477$ at 96, 120, 144, and 168 hours.
F1 declines monotonically with return period at every lead.

\begin{table}[h]
\centering
\caption{RiverMamba F1-score for floods at 4--7-day leads on the CONUS land mask, 2024 test window.
Hapi is not evaluated at these leads.
CONUS-pooled from the per-subset TP/FP/FN of the same evaluation used throughout the paper.}
\label{tab:rm_extended}
\small
\setlength{\tabcolsep}{4pt}
\begin{tabular}{@{}lccccccccc@{}}
\toprule
& \multicolumn{9}{c}{\textbf{Return period (years)}} \\
\cmidrule(lr){2-10}
\textbf{Lead} & 1.5 & 2 & 5 & 10 & 20 & 50 & 100 & 200 & 500 \\
\midrule
96\,h (4\,d)  & 0.580 & 0.542 & 0.352 & 0.224 & 0.146 & 0.078 & 0.049 & 0.030 & 0.017 \\
120\,h (5\,d) & 0.566 & 0.518 & 0.328 & 0.200 & 0.124 & 0.063 & 0.036 & 0.022 & 0.013 \\
144\,h (6\,d) & 0.571 & 0.490 & 0.283 & 0.160 & 0.090 & 0.041 & 0.024 & 0.014 & 0.009 \\
168\,h (7\,d) & 0.477 & 0.458 & 0.268 & 0.151 & 0.084 & 0.037 & 0.023 & 0.016 & 0.010 \\
\bottomrule
\end{tabular}
\end{table}

\section{Flood Forecasting at Extreme Return Periods}
\label{app:extreme_tail}

The GloFAS v4.0 return levels are obtained from a Gumbel fit to a 44-year record of annual maxima from 1979--2022, as described in Section~\ref{sec:thresholds}.
The 100-yr level therefore extrapolates to $2.3\times$ the record length, which is routine in design-flood practice.
The 200- and 500-yr levels reach $4.5\times$ and $11\times$ the fitted record length and require substantially longer extrapolation.
The main-text F1 results in Tables~\ref{tab:flood_f1} and~\ref{tab:ablation} are therefore restricted to return periods $\leq 100$~yr.
Table~\ref{tab:extreme_tail} provides CONUS-aggregated results at the more uncertain 200- and 500-yr levels for all four sources at 24, 48, and 72~h leads.
Persistence leads both extreme rows at 24-h lead, driven by the tiny event count per cell (median $\leq 2$ over the test window on the CONUS land mask).
Hapi leads at both levels at 48- and 72-h leads, although the difference at the 500-yr level and 48~h is only $0.001$.
Hapi outperforms GloFAS at every lead.
We report these extreme-tail rankings descriptively without tests of statistical significance.

\begin{table}[h]
\centering
\caption{CONUS-aggregated F1-score for floods at the 200- and 500-yr return periods.
These levels extrapolate the GloFAS Gumbel fit to $4.5\times$ and $11\times$ its 44-year record. The rankings are descriptive and are not accompanied by significance tests.}
\label{tab:extreme_tail}
\small
\setlength{\tabcolsep}{5pt}
\begin{tabular}{@{}llcc@{}}
\toprule
\textbf{Model} & \textbf{Lead} & \textbf{200-yr F1} & \textbf{500-yr F1} \\
\midrule
persistence  & 24\,h & \textbf{0.459} & \textbf{0.431} \\
GloFAS       & 24\,h & 0.234 & 0.196 \\
RiverMamba   & 24\,h & 0.114 & 0.086 \\
\hapirow Hapi & 24\,h & 0.429 & 0.367 \\
\midrule
persistence  & 48\,h & 0.199 & 0.149 \\
GloFAS       & 48\,h & 0.154 & 0.117 \\
RiverMamba   & 48\,h & 0.069 & 0.047 \\
\hapirow Hapi & 48\,h & \textbf{0.219} & \textbf{0.150} \\
\midrule
persistence  & 72\,h & 0.110 & 0.076 \\
GloFAS       & 72\,h & 0.101 & 0.068 \\
RiverMamba   & 72\,h & 0.046 & 0.030 \\
\hapirow Hapi & 72\,h & \textbf{0.153} & \textbf{0.101} \\
\bottomrule
\end{tabular}
\end{table}

\section{Per-Subset Flood Forecasting Skill}
\label{app:splits}

The per-lead grids extend the per-subset comparison in Section~\ref{sec:results_subset}.
Table~\ref{tab:splits_f1} in the main text reports F1-scores for floods at all seven primary return periods, and Table~\ref{tab:splits_cont} below reports continuous discharge skill, both for every subset at the nominal 24-, 48-, and 72-h leads reported by each system.

Both models' pooled F1 aggregates TP, FP, and FN across the cells of a subset before the ratio is formed.
Hapi's numbers come from the per-lead \texttt{daily\_discharge.nc}, and RiverMamba's come from its per-lead evaluation logs.
The AIFAS subset contains 146{,}265 cells co-located with operational AIFAS diagnostic river points.
The flood-active subset contains 135{,}523 cells exceeding the 2-yr threshold at least once in 2024.
Upstream drainage area partitions all 414{,}795 valid CONUS cells into small rivers below 500~km$^2$, medium rivers from 500 to below 5{,}000~km$^2$, and large rivers of at least 5{,}000~km$^2$, containing 338{,}893, 51{,}744, and 24{,}158 cells, respectively.
The AIFAS and flood-active subsets overlap these drainage groups.

\begin{table}[h]
\centering
\caption{Per-subset continuous discharge skill, pooled over paired cell--day values within each subset, at nominal forecast leads over CONUS in the 2024 test window. MAE is in m$^3$\,s$^{-1}$. NSE and KGE are dimensionless. Bold marks lower MAE or higher NSE and KGE for each subset and lead, using values before rounding to two decimal places.}
\label{tab:splits_cont}
\small
\setlength{\tabcolsep}{4pt}
\begin{tabular}{@{}ll ccc ccc ccc@{}}
\toprule
& & \multicolumn{3}{c}{\textbf{24\,h}} & \multicolumn{3}{c}{\textbf{48\,h}} & \multicolumn{3}{c}{\textbf{72\,h}} \\
\cmidrule(lr){3-5}\cmidrule(lr){6-8}\cmidrule(lr){9-11}
\textbf{Subset} & \textbf{Model} & MAE & NSE & KGE & MAE & NSE & KGE & MAE & NSE & KGE \\
\midrule
\hapirow \multirow{2}{*}{AIFAS} & Hapi & \textbf{4.14} & \textbf{1.00} & 0.99 & \textbf{7.84} & 0.99 & 0.98 & 10.91 & 0.98 & 0.98 \\
                    & RM   & 5.39 & 1.00 & \textbf{1.00} & 7.89 & 0.99 & \textbf{0.99} & \textbf{9.91} & \textbf{0.99} & \textbf{0.99} \\
\midrule
\hapirow \multirow{2}{*}{flood-active} & Hapi & \textbf{1.56} & \textbf{0.99} & 0.97 & \textbf{2.77} & \textbf{0.97} & 0.95 & \textbf{3.67} & \textbf{0.96} & 0.93 \\
                    & RM   & 2.90 & 0.98 & \textbf{0.99} & 4.10 & 0.95 & \textbf{0.97} & 4.83 & 0.94 & \textbf{0.97} \\
\midrule
\hapirow \multirow{2}{*}{Small rivers} & Hapi & \textbf{0.10} & \textbf{0.82} & \textbf{0.82} & \textbf{0.13} & \textbf{0.78} & \textbf{0.81} & \textbf{0.15} & \textbf{0.76} & \textbf{0.80} \\
                    & RM   & 0.22 & 0.60 & 0.80 & 0.24 & 0.54 & 0.77 & 0.26 & 0.58 & 0.79 \\
\midrule
\hapirow \multirow{2}{*}{Medium rivers} & Hapi & \textbf{1.14} & \textbf{0.91} & \textbf{0.90} & \textbf{1.72} & \textbf{0.86} & \textbf{0.87} & \textbf{2.09} & \textbf{0.84} & \textbf{0.87} \\
                    & RM   & 2.69 & 0.76 & 0.87 & 3.32 & 0.65 & 0.82 & 3.65 & 0.68 & 0.84 \\
\midrule
\hapirow \multirow{2}{*}{Large rivers} & Hapi & \textbf{23.15} & \textbf{1.00} & 0.99 & 45.05 & 0.99 & 0.98 & 63.52 & 0.98 & 0.98 \\
                    & RM   & 27.45 & 1.00 & \textbf{1.00} & \textbf{41.84} & \textbf{0.99} & \textbf{0.99} & \textbf{53.85} & \textbf{0.99} & \textbf{0.99} \\
\bottomrule
\end{tabular}
\end{table}

\begin{figure*}[!htbp]
    \centering
    \includegraphics[width=\textwidth]{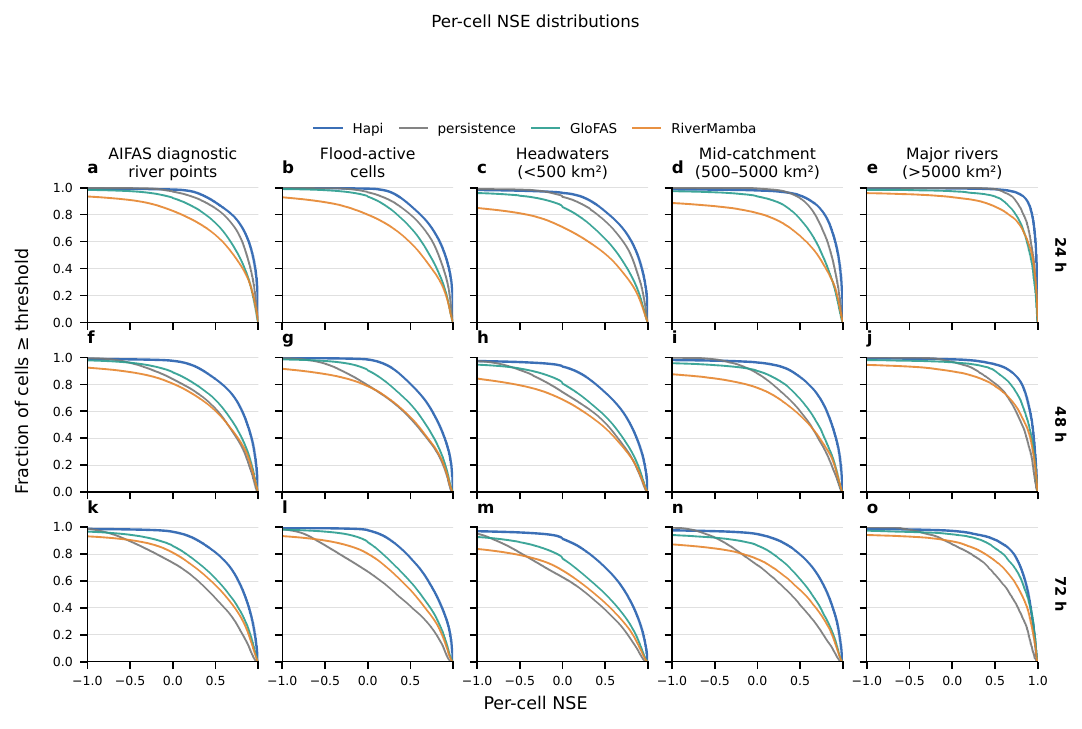}
    \caption{Complementary empirical distributions of per-cell NSE for four forecast sources across five spatial subsets (columns) and nominal 24-, 48-, and 72-h leads (rows).
    Each curve gives the fraction of cells attaining at least the indicated NSE and uses every finite cell value.
    The horizontal axis is displayed from $-1$ to 1. Values below $-1$ remain in the empirical sample and contribute to the fraction below the plotted range.}
    \label{fig:ecdf_nse}
\end{figure*}

\begin{figure*}[!htbp]
    \centering
    \includegraphics[width=\textwidth]{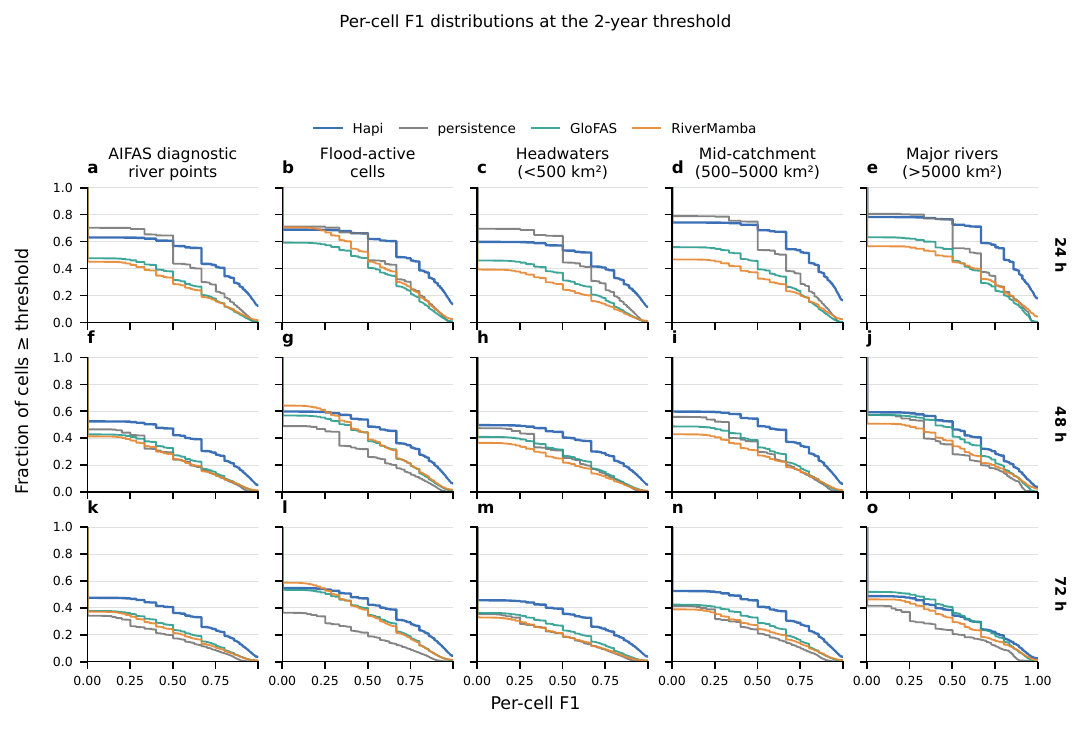}
    \caption{Complementary empirical distributions of per-cell F1 at the 2-yr return-period threshold for four forecast sources across five spatial subsets (columns) and nominal 24-, 48-, and 72-h leads (rows).
    Each curve gives the fraction of event-containing cells attaining at least the indicated F1 and uses every finite cell value.
    Hapi and RiverMamba are compared at their nominal reported leads.}
    \label{fig:ecdf_f1}
\end{figure*}

\textbf{Per-cell distributions.} The pooled F1 numbers in Table~\ref{tab:splits_f1} aggregate TP, FP, and FN over every cell in a subset and report a single number, which the heaviest-event cells dominate.
Figures~\ref{fig:ecdf_nse} and~\ref{fig:ecdf_f1} show the corresponding per-cell distributions for Hapi, persistence, GloFAS, and RiverMamba.
At each skill threshold, the vertical axis gives the fraction of cells attaining that value or higher.
The NSE figures display thresholds from $-1$ to 1 without clipping the underlying values. The F1 figures use the 2-yr return-period threshold and cells with $\geq$1 event over the test window.
The curves use every finite cell value and compare Hapi and RiverMamba at their nominal leads. RiverMamba provides leads 1--7, and the figures show leads 1--3.
The figures report per-cell distributions, whereas Table~\ref{tab:splits_f1} reports ratios pooled across cells.

The pattern is consistent across all three subset groups: Hapi's curve generally lies above the three baselines and declines least over 24--72~h.
Persistence declines most rapidly, with GloFAS and RiverMamba intermediate over most of the skill range.
Per-cell F1 retains a visible mass at 0 (cells where the model never aligns with a target event) and at the small-integer rationals $\{1/3, 1/2, 2/3, 1\}$ (cells with 1--3 events where partial agreement is possible).
The steps are intrinsic to event sparsity at fine spatial resolution.

Per-cell F1 is sensitive to timing at this resolution: for a single-event cell, a predicted exceedance displaced by one day gives F1$=0$, since TP$=0$ and FP=FN=$1$.
Most cells at the 2-yr threshold see only one or two events over the test window, so the initial drop at F1$=0$ reflects timing sensitivity at sparse event counts.
The resulting mass at zero occurs for all four sources and does not by itself imply complete model failure.
The pooled F1 in Table~\ref{tab:splits_f1} avoids this floor because TP, FP, and FN are accumulated across cells before the F1 ratio is computed.
Per-cell F1 at the other eight return periods (1.5, 5, 10, 20, 50, 100, 200, 500\,yr) is in the supplementary file set and preserves the same model ordering.

\clearpage
\section{Per-Cell Continuous Skill}
\label{app:percell}

The continuous metrics in Table~\ref{tab:main_discharge} are pooled over space and time, using all paired cell--day values in each calculation.
On a domain spanning five orders of magnitude of discharge, pooling gives high-discharge rivers greater influence, particularly on RMSE and NSE through their squared-error terms.
Typical-cell skill is quantified by computing NSE, KGE, and the KGE bias ratio $\beta$ independently at every cell and taking the median, as reported in Table~\ref{tab:percell}.
A cell is included when all three sources are defined at that lead and the cell lies inside the CONUS mask: 374{,}297 of 414{,}795 cells, 90.2\%.
The excluded 9.8\% have near-zero target variance over the test window, which makes the NSE denominator collapse.

Pooled, GloFAS overtakes Hapi from 48-h lead onward on NSE and KGE.
Per cell, Hapi has the highest median NSE at every lead (0.902 / 0.823 / 0.776 against 0.698 / 0.637 / 0.559 for GloFAS) and the highest median KGE at 48 and 72~h.
At 72-h lead Hapi's per-cell NSE exceeds GloFAS's at 79.2\% of cells with a median difference of $+0.142$, and persistence at 92.3\% of cells with a median difference of $+0.363$.
Pooling conceals a broad advantage at typical cells behind the behavior of a small number of very large ones.
The pooled metrics characterize skill over all cell--day values, whereas the per-cell medians characterize skill at a typical location.

The per-cell median $\beta$ is 0.973, 0.967 and 0.962 at 24, 48 and 72~h, indicating slight underprediction at a typical cell despite the pooled result.
Persistence has a median bias ratio of $\beta = 1.000$ to three decimal places at every lead.

\begin{figure}[!htbp]
    \centering
    \includegraphics[width=\linewidth]{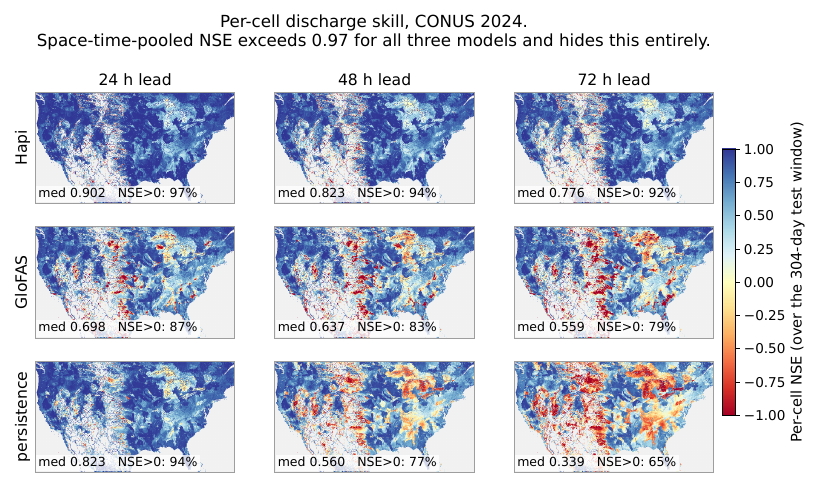}
    \caption{Per-cell NSE over the 2024 test window for each model and lead, CONUS land mask.
    Cells where the score is undefined (near-zero target variance) are left blank.
    The pooled values in Table~\ref{tab:main_discharge} are dominated by high-discharge cells and do not reflect the typical cell shown here.}
    \label{fig:percell_maps}
\end{figure}

\begin{figure}[!htbp]
    \centering
    \includegraphics[width=\linewidth]{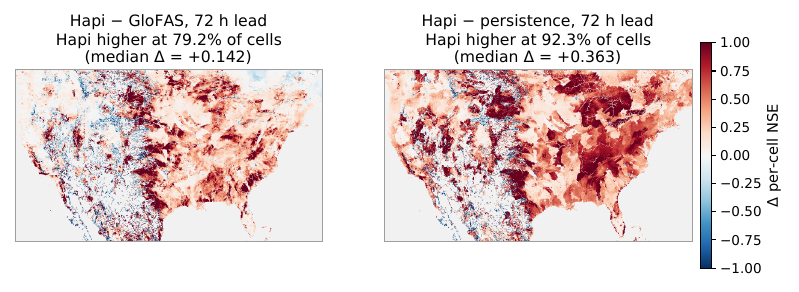}
    \caption{Per-cell NSE difference, Hapi minus GloFAS, by lead.
    Positive (blue) favors Hapi.
    The remaining GloFAS-favored cells form coherent regional clusters. Possible explanations of these regional differences are discussed in Section~\ref{sec:discussion}.}
    \label{fig:percell_diff}
\end{figure}

Figure~\ref{fig:percell_maps} maps the per-cell scores and Figure~\ref{fig:percell_diff} the Hapi-minus-GloFAS difference.
The difference map shows that the residual GloFAS advantage clusters in the upper Midwest, the Northeast and parts of the mountain West.

\FloatBarrier
\section{Paired Block Bootstrap and Multiplicity Correction}
\label{app:bootstrap_paired}

The F1-scores for floods in Table~\ref{tab:flood_f1} come from one 2024 test window whose daily values are strongly autocorrelated, so any statement about which model ranks higher needs an uncertainty estimate attached to it.
We use a moving-block bootstrap~\citep{kunsch1989jackknife} over the time axis.

\paragraph{Resampling scheme.}
Within each of $R = 10{,}000$ replicates we draw $\lceil n/L \rceil$ blocks of $L = 30$ consecutive days with replacement, truncate to $n$ days, and recompute the pooled contingency table.
% Pairing is performed separately at each nominal lead: Hapi--persistence and Hapi--GloFAS use 302 valid dates, whereas Hapi--RiverMamba uses their 294 common valid dates.
The same block draw is applied to every model within a replicate, making the difference between two models the resampled quantity.
Whole days are resampled, preserving within-day spatial dependence without treating the $414{,}795$ spatially clustered cells as independent observations.
Within each replicate, F1 and SEDI are recomputed from the pooled resampled contingency counts.

\paragraph{Paired differences and marginal intervals.}
Every model is scored on the same days, and their errors are strongly positively correlated because all models encounter the same sequence of hydrological conditions.
Pairing accounts for shared day-to-day variation and reduces the variance of the difference relative to treating the model estimates as independent.
Requiring the two marginal 95\% intervals to be disjoint is more conservative than testing the paired difference directly.
For the persistence and GloFAS comparisons, the marginal intervals overlap in every 72-h cell, whereas the paired test separates 20 of 28 cells.
The paired test is therefore the appropriate inference for these correlated forecasts.

\paragraph{Block length.}
The block must be long enough to preserve temporal dependence in the resampled series, which here is the daily count of cells exceeding the threshold.
This dependence decreases sharply as return period increases.
Frequent exceedances often occur when the catchment has been wet for a long time.
Rare exceedances usually occur during single storms.
The observed series' lag-1 autocorrelation and integrated autocorrelation time $\tau_{\mathrm{int}} = 1 + 2\sum_k \rho_k$, truncated at the first non-positive $\rho_k$, appear in Table~\ref{tab:block_dependence}.

\begin{table}[!htbp]
\centering
\caption{Temporal dependence of the observed daily exceedance count on the CONUS mask, 2024 test window, by return-period threshold.
Computed at 24-h lead. The series is the common verification target, so the other leads agree to within four cells per day.
$\tau_{\mathrm{int}}$ is in days.}
\label{tab:block_dependence}
\small
\begin{tabular}{@{}lrrrrrrr@{}}
\toprule
\textbf{Return period (yr)} & 1.5 & 2 & 5 & 10 & 20 & 50 & 100 \\
\midrule
Exceeding cells per day     & 7{,}883 & 3{,}453 & 689 & 270 & 126 & 59 & 36 \\
Lag-1 autocorrelation       & 0.938 & 0.908 & 0.847 & 0.747 & 0.652 & 0.614 & 0.594 \\
$\tau_{\mathrm{int}}$ (d)   & 42.3 & 25.7 & 10.0 & 4.8 & 3.5 & 3.0 & 2.7 \\
\bottomrule
\end{tabular}
\end{table}

No single block length matches all seven columns.
We set $L = 30$~days near the upper end of the estimated dependence range and assess sensitivity to this choice below.
At this block length, the 50- and 100-yr F1 and SEDI differences at 72-h lead against persistence and GloFAS are not significant after Holm correction. Their nonsignificance does not establish equivalence.

\paragraph{Sensitivity to the block length.}
Re-running the whole procedure at block lengths from 7 to 60~days leaves the conclusion in place, as shown in Table~\ref{tab:block_sensitivity}.
The surviving count moves over a narrow range with no trend across 7--45~days.
At a block length of 60~days, each replicate draws five blocks for the 294-day comparisons and six blocks for the 302-day comparisons, with the final block truncated as needed.

\begin{table}[!htbp]
\centering
\caption{Cells surviving Holm correction (of 21 per family) as the bootstrap block length varies, $R = 10{,}000$ throughout.
The 30-day column corresponds to the block length used in Figure~\ref{fig:bootstrap_paired}.
Reproduced by \texttt{bootstrap\_block\_sensitivity.py}, which also emits Table~\ref{tab:block_dependence}.}
\label{tab:block_sensitivity}
\small
\begin{tabular}{@{}lcccccc@{}}
\toprule
\textbf{Block length} & 7\,d & 15\,d & 21\,d & \textbf{30\,d} & 45\,d & 60\,d \\
\midrule
F1, Hapi\,--\,persistence   & 14 & 14 & 14 & \textbf{14} & 12 & 13 \\
F1, Hapi\,--\,GloFAS        & 17 & 17 & 17 & \textbf{18} & 17 & 21 \\
F1, Hapi\,--\,RiverMamba    & 21 & 21 & 21 & \textbf{21} & 21 & 21 \\
SEDI, Hapi\,--\,persistence & 12 & 12 & 12 & \textbf{12} & 12 & 12 \\
SEDI, Hapi\,--\,GloFAS      & 17 & 16 & 16 & \textbf{17} & 16 & 19 \\
SEDI, Hapi\,--\,RiverMamba  & 7 & 7 & 6 & \textbf{6} & 6 & 6 \\
\midrule
Total (of 126)              & 88 & 87 & 86 & \textbf{88} & 84 & 92 \\
\bottomrule
\end{tabular}
\end{table}

\paragraph{$p$-values and multiplicity.}
Each cell's $p$-value is the two-sided percentile bootstrap $p$: the proportion of replicates whose difference falls on the opposite side of zero from the point estimate, doubled, with the $(1+k)/(R+1)$ correction of \citet{davison1997bootstrap} so that no cell can report $p = 0$.
The smallest attainable raw value is therefore $2/(R+1)$.
Because 21 (lead $\times$ return-period) cells are inspected at once, we apply a Holm step-down correction~\citep{holm1979simple} across each 21-cell family, separately per comparison and per metric, for six families and 126 cells in total.
We define six families by crossing the two prespecified metrics (F1 and SEDI) with the three model comparisons (Hapi--persistence, Hapi--GloFAS, and Hapi--RiverMamba).
Within each family, Holm correction covers all leads and return periods.
Holm holds under arbitrary dependence between the tests, which is what this application needs: the seven return periods within a lead are nested, since every 100-yr exceedance is also a 1.5-yr exceedance.
After correction the smallest attainable adjusted value is $42/(R+1)$.
At $R = 1{,}000$, this bound is $0.042$, placing a comparison with no sign reversals just below $\alpha = 0.05$ because of Monte Carlo resolution. Increasing $R$ to $10{,}000$ reduces the bound to $0.0042$.
Increasing $R$ from 1{,}000 to 10{,}000 changed few uncorrected decisions but lowered the minimum attainable adjusted $p$-value.

\begin{figure}[!htbp]
    \centering
    \includegraphics[width=\linewidth]{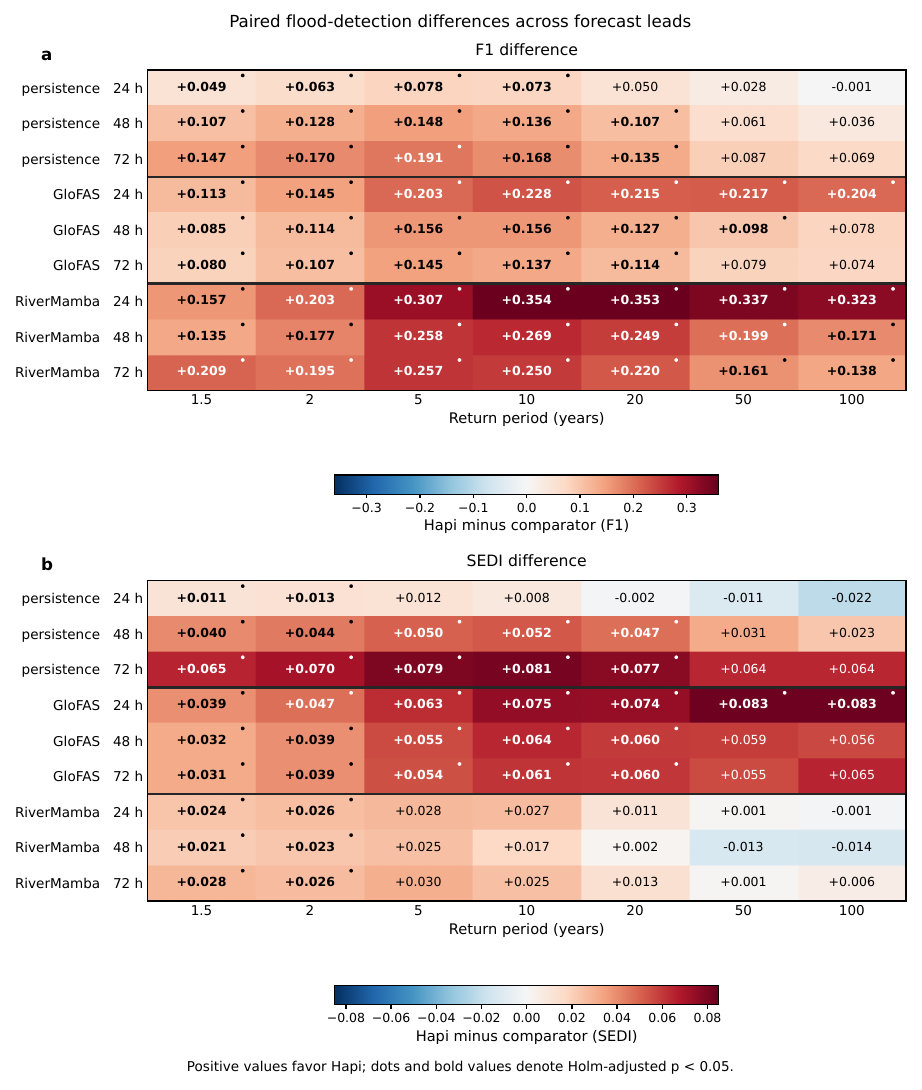}
    \caption{Paired moving-block-bootstrap differences in flood forecasting skill over the 2024 CONUS test window.
    Values are Hapi minus the indicated comparator at each nominal forecast lead and return-period threshold. Positive values favor Hapi.
    Panel a shows F1 differences and panel b SEDI differences.
    Bold values and dots indicate $p<0.05$ after Holm correction across the 21 lead--return-period cells within each comparison and metric.
    % Hapi--persistence and Hapi--GloFAS use 302 valid days per lead. Hapi--RiverMamba uses the 294 dates available for both models.
    RiverMamba's latest discharge input is one day older than Hapi's at the same nominal lead.
    The corresponding 2.5th--97.5th percentile intervals are retained in the source-data file.}
    \label{fig:bootstrap_paired}
\end{figure}

\paragraph{Primary model-comparison results.}
The underlying bootstrap output also reports the paired point difference and its 2.5th--97.5th percentile interval for every cell. The intervals quantify effect-size uncertainty, whereas Figure~\ref{fig:bootstrap_paired} emphasizes the point differences and multiplicity-adjusted decisions.
At 24-h lead Hapi's F1 advantage over the GloFAS forecast is significant after correction at \emph{every} return period from 1.5 to 100~yr.
Hapi's F1 advantage over RiverMamba survives correction in all 21 cells. The corresponding SEDI difference survives in six cells, at the 1.5- and 2-yr thresholds at each lead.
Across all 126 cells, 88 survive correction, distributed 29 / 30 / 29 over the 24-, 48- and 72-h leads.
None of the 126 comparisons shows a significant loss for Hapi after Holm correction. The seven negative point differences comprise six SEDI comparisons and one F1 comparison.

% \paragraph{Inference scope.}
% With 30-day blocks, the 294- and 302-day paired windows provide roughly ten effective temporal samples. A multi-year evaluation would increase precision.

\clearpage
\section{Base-Rate-Invariant Skill and the Precision--Recall Decomposition}
\label{app:sedi}

F1 depends on the event base rate, which falls by orders of magnitude as the return-period threshold rises.
Consequently, F1 supports comparisons among models at a fixed threshold but not comparisons of skill across thresholds.
For cross-threshold comparisons, we use SEDI because it remains non-degenerate as the event base rate tends to zero.

\begin{figure}[!htbp]
    \centering
    \includegraphics[width=\linewidth]{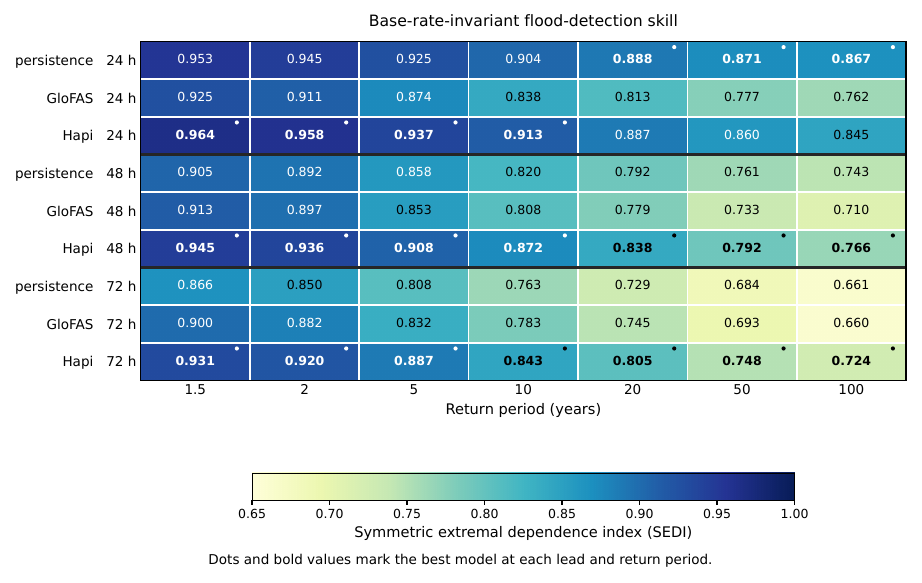}
    \caption{Symmetric extremal dependence index (SEDI) for floods on the CONUS land mask over the 2024 test window.
    SEDI is invariant to the event base rate and ranges from $-1$ to 1, with 0 indicating no skill.
    Rows group persistence, GloFAS, and Hapi at 24-, 48-, and 72-h leads. Columns give the GloFAS flood thresholds for different return periods.
    Exact SEDI values are printed in each cell.
    Bold values and dots mark the best model within each lead--return-period comparison.}
    \label{fig:sedi}
\end{figure}

Figure~\ref{fig:sedi} reports SEDI at the same 21 lead--threshold combinations as F1.
Hapi's 72-h SEDI decreases from 0.931 at the 1.5-yr threshold to 0.724 at the 100-yr threshold, indicating lower skill for rarer events under this base-rate-invariant measure.
Among Hapi, GloFAS, and persistence, Hapi has the highest SEDI in 18 of the 21 comparisons, including all seven at both 48- and 72-h leads.
The three exceptions are all at 24-h lead, at the 20-, 50-, and 100-yr thresholds, where persistence leads.

\begin{figure}[!htbp]
    \centering
    \includegraphics[width=\linewidth]{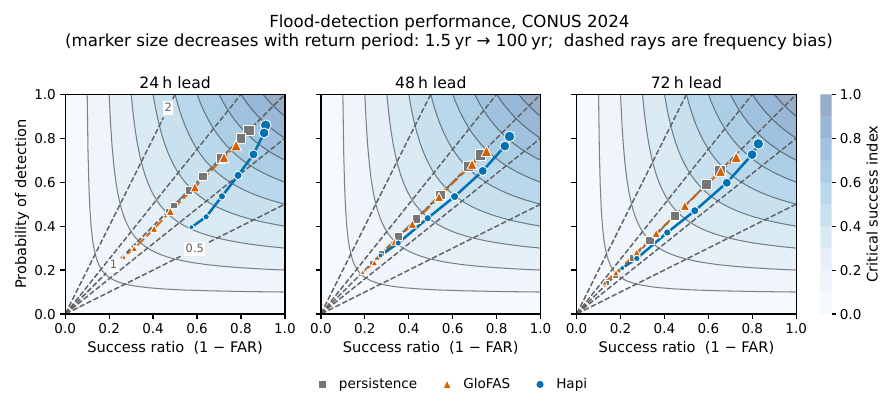}
    \caption{Performance diagram for flood forecasting over the CONUS land mask in the 2024 test window.
    The horizontal axis shows success ratio (precision), and the vertical axis shows probability of detection (recall). Performance improves toward the upper-right corner, and the shaded contours show CSI.
    Dashed rays show frequency bias, with values above and below one indicating overforecasting and underforecasting, respectively.
    Marker size decreases from the 1.5-yr to the 100-yr return-period threshold.}
    \label{fig:performance_diagram}
\end{figure}

To separate the contributions of missed events and false alarms, we also report precision, recall, and CSI in Table~\ref{tab:precision_recall_csi}.
Figure~\ref{fig:performance_diagram} shows their geometric relationship.
Figure~\ref{fig:performance_diagram} favors Hapi most clearly at 48- and 72-h leads, where it lies above and to the right of both baselines at every threshold and therefore occupies higher CSI contours.
At 24~h, Hapi retains higher precision, but persistence has higher recall at the rarest thresholds.
Hapi's advantage over the GloFAS forecast at 72-h lead is larger in precision than in recall at every return period: at the 5-yr threshold it gains $+0.191$ in precision and $+0.106$ in recall, and at the 100-yr threshold $+0.081$ and $+0.067$, respectively.
Hapi has higher recall than GloFAS at every lead--threshold combination.

At 72-h lead and the 1.5-yr threshold, its precision exceeds GloFAS by $0.100$, its recall by $0.061$, and its CSI by $0.104$. Hapi also has higher CSI at 48~h ($0.713$ versus $0.597$).
GloFAS does not exceed Hapi on CSI in any comparison in Table~\ref{tab:precision_recall_csi}.
Persistence does so in one cell only, at 24-h lead and the 100-yr threshold, and by $0.001$ ($0.307$ against $0.306$), consistent with its known short-lead advantage.
Hapi's precision exceeds its recall in every cell but one, corresponding to fewer predicted than observed exceedances. The exception is 72-h lead at the 100-yr threshold, where recall edges ahead by $0.001$ ($0.209$ against $0.208$).
Persistence's precision and recall differ by at most $0.001$ over the test window.

\begin{table}[!htbp]
\centering
\caption{Precision, recall, and critical success index (CSI) for floods at GloFAS return periods, CONUS land mask, 2024 test window. Metric definitions are given in Appendix~\ref{app:metric_definitions}. Recall exceeding precision indicates overforecasting of floods. The reverse indicates underforecasting. Bold marks the highest value for each lead and return period.}
\label{tab:precision_recall_csi}
\small
\setlength{\tabcolsep}{3pt}
\begin{tabular}{@{}llccccccc@{}}
\toprule
 & & \multicolumn{7}{c}{\textbf{Return period (years)}} \\
\cmidrule(lr){3-9}
\textbf{Model} & \textbf{Lead} & 1.5 & 2 & 5 & 10 & 20 & 50 & 100 \\
\midrule
\multicolumn{9}{@{}l}{\textit{Precision}} \\
persistence & 24\,h & 0.836 & 0.800 & 0.709 & 0.627 & 0.562 & 0.496 & 0.470 \\
persistence & 48\,h & 0.726 & 0.672 & 0.544 & 0.434 & 0.353 & 0.275 & 0.233 \\
persistence & 72\,h & 0.653 & 0.590 & 0.447 & 0.334 & 0.256 & 0.177 & 0.140 \\
GloFAS & 24\,h & 0.778 & 0.723 & 0.590 & 0.478 & 0.405 & 0.314 & 0.267 \\
GloFAS & 48\,h & 0.754 & 0.688 & 0.538 & 0.417 & 0.336 & 0.243 & 0.190 \\
GloFAS & 72\,h & 0.727 & 0.656 & 0.493 & 0.362 & 0.274 & 0.179 & 0.127 \\
\hapirow Hapi & 24\,h & \textbf{0.912} & \textbf{0.905} & \textbf{0.857} & \textbf{0.787} & \textbf{0.713} & \textbf{0.641} & \textbf{0.574} \\
\hapirow Hapi & 48\,h & \textbf{0.859} & \textbf{0.837} & \textbf{0.737} & \textbf{0.609} & \textbf{0.486} & \textbf{0.352} & \textbf{0.272} \\
\hapirow Hapi & 72\,h & \textbf{0.827} & \textbf{0.799} & \textbf{0.684} & \textbf{0.538} & \textbf{0.413} & \textbf{0.274} & \textbf{0.208} \\
\midrule
\multicolumn{9}{@{}l}{\textit{Recall}} \\
persistence & 24\,h & 0.836 & 0.799 & 0.708 & 0.627 & \textbf{0.562} & \textbf{0.496} & \textbf{0.470} \\
persistence & 48\,h & 0.726 & 0.671 & 0.544 & 0.434 & 0.353 & 0.275 & 0.233 \\
persistence & 72\,h & 0.653 & 0.590 & 0.447 & 0.334 & 0.256 & 0.177 & 0.140 \\
GloFAS & 24\,h & 0.767 & 0.713 & 0.577 & 0.467 & 0.388 & 0.299 & 0.261 \\
GloFAS & 48\,h & 0.741 & 0.682 & 0.533 & 0.410 & 0.330 & 0.234 & 0.191 \\
GloFAS & 72\,h & 0.714 & 0.650 & 0.492 & 0.367 & 0.280 & 0.188 & 0.142 \\
\hapirow Hapi & 24\,h & \textbf{0.859} & \textbf{0.824} & \textbf{0.727} & \textbf{0.631} & 0.535 & 0.442 & 0.395 \\
\hapirow Hapi & 48\,h & \textbf{0.808} & \textbf{0.764} & \textbf{0.651} & \textbf{0.535} & \textbf{0.436} & \textbf{0.323} & \textbf{0.266} \\
\hapirow Hapi & 72\,h & \textbf{0.775} & \textbf{0.726} & \textbf{0.598} & \textbf{0.470} & \textbf{0.371} & \textbf{0.253} & \textbf{0.209} \\
\midrule
\multicolumn{9}{@{}l}{\textit{CSI}} \\
persistence & 24\,h & 0.719 & 0.666 & 0.549 & 0.457 & 0.391 & 0.329 & \textbf{0.307} \\
persistence & 48\,h & 0.570 & 0.505 & 0.373 & 0.277 & 0.214 & 0.160 & 0.132 \\
persistence & 72\,h & 0.485 & 0.418 & 0.288 & 0.201 & 0.147 & 0.097 & 0.075 \\
GloFAS & 24\,h & 0.629 & 0.560 & 0.412 & 0.310 & 0.247 & 0.181 & 0.152 \\
GloFAS & 48\,h & 0.597 & 0.521 & 0.365 & 0.260 & 0.199 & 0.136 & 0.105 \\
GloFAS & 72\,h & 0.563 & 0.485 & 0.327 & 0.223 & 0.161 & 0.101 & 0.072 \\
\hapirow Hapi & 24\,h & \textbf{0.794} & \textbf{0.759} & \textbf{0.648} & \textbf{0.539} & \textbf{0.441} & \textbf{0.355} & 0.306 \\
\hapirow Hapi & 48\,h & \textbf{0.713} & \textbf{0.665} & \textbf{0.528} & \textbf{0.398} & \textbf{0.298} & \textbf{0.202} & \textbf{0.155} \\
\hapirow Hapi & 72\,h & \textbf{0.667} & \textbf{0.613} & \textbf{0.468} & \textbf{0.335} & \textbf{0.243} & \textbf{0.151} & \textbf{0.116} \\
\bottomrule
\end{tabular}
\end{table}

\end{document}